\documentclass[12pt]{article}
\usepackage{graphicx}
\usepackage{tabularx}
\usepackage{scalerel}
\usepackage{graphicx}
\usepackage{amsmath, amssymb}
\usepackage{blindtext}
\usepackage{subcaption}

\usepackage{caption}
\usepackage{tabu}
\usepackage{apacite}
\usepackage{natbib}
\usepackage{setspace}
\usepackage{hyperref}
\usepackage{lscape}
\usepackage{colortbl,hhline}
\usepackage{hyperref}
\usepackage{times}
\usepackage{setspace}
\usepackage{chngcntr}
\usepackage{booktabs,multirow}

\numberwithin{figure}{section}
\numberwithin{equation}{section}
\counterwithin{table}{section}
\usepackage{times}
\newenvironment{sciabstract}{%
	\begin{quote} \bf}
	{\end{quote}}
\title{Robust non-parametric mortality and fertility modelling and forecasting: Gaussian process regression approaches} 
\author
{Ka Kin Lam\footnote{Corresponding author. E-mail: kkl21@leicester.ac.uk (K.K.Lam).}, Bo Wang\\
	\\
	\normalsize{School of Mathematics and Actuarial Science, University of Leicester,}\\
	\normalsize{Leicester, LE1 7RH, UK}\\
}

\date{}
\begin{document} 
	\baselineskip24pt
	\maketitle 
\begin{sciabstract}
	\begin{center}
		Abstract
	\end{center}
	\normalfont A rapid decline in mortality and fertility has become major issues in many developed countries over the past few decades.  A precise model for forecasting demographic movements is important for decision making in social welfare policies and resource budgeting among the government and many industry sectors. This article introduces a novel non-parametric approach using Gaussian process regression with a natural cubic spline mean function and a spectral mixture covariance function for mortality and fertility modelling and forecasting. Unlike most of the existing approaches in  demographic modelling literature, which rely on time parameters to decide the movements of the whole mortality or fertility curve shifting from one year to another over time, we consider the mortality and fertility curves from their components of all age-specific mortality and fertility rates and assume each of them following a Gaussian process over time to fit the whole curves in a discrete but intensive style. The proposed Gaussian process regression approach shows significant improvements in terms of preciseness and robustness compared to other mainstream demographic modelling approaches in the short-, mid- and long-term forecasting using the mortality and fertility data of several developed countries in our numerical experiments. 
\end{sciabstract}
\section{Introduction}\label{sec: GPR Introduction}
There has been an increasing demand for demographic modelling and forecasting over the last few decades, driven by many developed countries are now suffering a rapid decline in mortality and fertility, leading to a significant increase in expenditures on health services for an ageing population and a shortage of future labour. A better understanding of the mortality and fertility patterns and trends is always of importance for all stakeholders in a society as the mortality forecasts, for example, play a vital role for the insurance and pensions industries in pricing their insurance products. The fertility predictions are also of great interest to the government and education sectors in planing children's welfare and educational services.
\par Unlike the biological and the medical methods, statisticians have developed very different and purely mathematical methods to model the demographic patterns and trends which are well-documented by \citet{preston2000demographymeasuring}. The history of demographic modelling with the mathematical approaches can be traced back to some deterministic models proposed in the mid-nineteenth century, see, for example, \citet{gompertz1825xxiv} and \citet{makeham1860law}. The deterministic models are, however, restricted with few fixed factors and have no stochastic process considered owing to the lack of computing capability in that early period.  With the advance of technology in computing, stochastic modelling has become a mainstream method of mortality and fertility curve fittings over the last three decades. A significant milestone of the demographic modelling literature is the seminal work done by \citet{lee1992modeling}, well-known as the Lee-Carter model, which can model and extrapolate a long-term mortality trend as a stochastic time series. It rapidly gained credit and popularity, given its simplicity and ability to capture most variations in demographic data evolved over time. For instance, \citet{lee1993modeling} applies it to fertility modelling and forecasting,  and there has also been a series of extensions, variants and modifications proposed afterwards, see, for example, \citet{bell1997comparing}, \citet{lee2001evaluating}, \citet{booth2002applying}, \citet{brouhns2002poisson}, \citet{renshaw2003lee}, \citet{cairns2006two} and \citet{hunt2014general}. 
\par In the meantime, non-parametric or data-driven techniques, which focus on the aspect of letting the data speak for themselves with no need to meet certain assumptions on parametric forms in the model calibration, have also been introduced and developed in demographic modelling. This kind of techniques can be dated back to the classical graduation technique \citep{whittaker1922new}, which maintains an agnostic view of historical experience and solely focuses on removing random fluctuations in observed data then directly extrapolates the past trend to the future. In more recent years, \citet{currie2004smoothing} use P-splines to smooth the historical mortality surface across both age and year dimensions before fitting. \citet{hyndman2007robust} extend the Lee-Carter model to a functional data framework with a non-parametric smoothing method that allows for smooth functions of age and is more robust for demographic modelling. Some other developments in this area include \citet{delwarde2007smoothing}, \citet{debon2010geostatistical}, \citet{li2015semiparametric}, \citet{ludkovski2018gaussian}, \citet{dokumentov2018bivariate}, \citet{wu2018gaussian} and \citet{alexopoulos2019bayesian}.
\par As an alternative to existing methods as well as having several desirable advantages over others, in this paper we propose a new non-parametric approach using Gaussian process regression with a natural cubic spline mean function and a spectral mixture covariance function for mortality and fertility modelling and forecasting. Unlike most of the existing approaches in demographic modelling, which depend on few parameters to decide the movement of the whole mortality or fertility curve shifting from one year to another, we consider the mortality and fertility curves from their components of all age-specific mortality and fertility rates and assume each of them following a Gaussian process over time to fit the whole curves in a discrete but dense style. 
\par More will be discussed in detail in the paper, and the rest of this paper is organised as follows. We briefly outline the theoretical background of Gaussian process regression in Section \ref{sec: GPR Theoretical background of Gaussian process regression}. In Section \ref{sec: GPR Methodology}, we describe the framework of the proposed Gaussian process regression model for mortality and fertility modelling and forecasting. We then illustrate the proposed model with applications to the empirical mortality and fertility data, followed by comparisons to other existing approaches in terms of the systematic differences and forecasting performances using the observed mortality and fertility data of ten developed countries in Section \ref{sec: Applications}. We lastly conclude this paper with discussions and remarks in Section \ref{sec: Discussion and conclusion remarks}. 
\section{Theoretical background of Gaussian process regression} \label{sec: GPR Theoretical background of Gaussian process regression}
Gaussian process regression (also known as Kriging) is a regression method which belongs to a class of Bayesian functional non-parametric approaches to inferencing and modelling of an unknown latent function. This approach can be seen as conditioning of test data by training data in a joint Gaussian distribution from a function space point of view. The theoretical basis for Gaussian process regression (GPR) was initially developed for estimating the most likely distribution of gold based on samples from a few boreholes in South Africa \citep{krige1951statistical}, and it was mainly used for spatial analysis and natural resources evaluation in geostatistics in former times, see, for example, \citet{matheron1963principles}, \citet{journel1978mining} and \citet{cressie1989geostatistics}.
\par Over the last two decades, GPR has gained its popularity rapidly as different statistical tools among the data science community, such as, \citet{engel2005reinforcement}, \citet{krause2008near} and \citet{neal2012bayesian}. It has also been treated as a powerful tool for regressions and forecasting problems in the field of economics thanks to its abilities to quantify the uncertainties associated with the historical experience of the observed data and generates the full stochastic trajectories for out-of-sample forecasts with prediction intervals under the Gaussian probabilistic framework. It is also capable of scaling to a large dataset with a minimum set of tunable (hyper-) parameters involved as well as capturing non-linear or periodic dynamics with a high degree of analytic tractability that extends the explorabilities and explainabilities of the classical linear regression model in more complicated scenarios.  Some economic forecasting applications with GPR can be found, for instance, \citet{alamaniotis2012evolutionary} perform a short-term load forecasting using an ensemble of GPR approaches and \citet{wu2012sparse} predict the tourism demand volume in Hong Kong using GPR. Since GPR is the fundamental technique applied in this article, we will give a brief review of GPR which will be used later. Readers can refer to \citet{mackay1998introduction} and \citet{williams2006gaussian} for more complete discussions of GPR.
\subsection{Gaussian process regression}
Let a training dataset with $n$ pairs of observations of univariate covariates and responses be $\lbrace t_{i}, y(t_{i})\rbrace^{n}_{i = 1}$. We consider the following non-linear regression model
\[
y(t_{i}) = f(t_{i}) + \epsilon(t_{i}), \hspace*{0.5cm} {\epsilon(t_{i})} \sim \mathcal{N}(0, \sigma^{2}),
\]
where $f(\cdot)$ is the unknown function that needs to be estimated, and $\{\epsilon(t_{i})\}_{i=1}^{n}$ are the independent and identically normally distributed error terms with mean 0 and constant variance $ \sigma^{2}$. 
\par Following the Gaussian process paradigm, the unknown function $\{f(t):t\in\mathcal{T}\}$ is assumed to have a Gaussian process prior with a specified mean function and a specified covariance function over the domain $\mathcal{T}$. It gives for all $t\in\mathcal{T}$
\begin{equation}\label{eq: GP prior distribtion}
	f(t) \sim \mathcal{GP}(\mu(t),K(t,t')),
\end{equation}
where  the mean function $\mu(t)$ and the covariance function $K(t,t')$ are defined as 
\[\mu(t) = \mathbb{E}[f(t)], \]
\[K(t,t') = \text{Cov}(f(t), f(t')) = \mathbb{E}[(f(t) - \mu(t))(f(t') - \mu(t'))].
\]
The specified mean function $\mu(\cdot)$ and the specified covariance function $K(\cdot,\cdot)$ in the Gaussian process prior reflect our prominent belief in the unknown function $f(\cdot)$,  prior to any information about the observed data. A well-specified prior mean function has a profound impact on the forecast performance since the mean function will dominate the forecast results of the Gaussian process regression model in regions far beyond from the historical data. Meanwhile, the specified covariance function encodes the correlation between any pair of outputs $\{y(t),y(t')\}$, which determines the relativeness of one point to another, such as smoothness or periodicity. 
\par A common covariance function example is the squared exponential covariance function which is used to model a smooth function. It has the form
\[
K(t,t') = h^{2}\text{exp}\bigg(-\frac{(t - t')^{2}}{2{l^{2}}}\bigg),
\]
where $h$ is the response-scale amplitude determining the variation of function values, $l$ is the characteristic length-scale which gives smooth variations in a covariate-scale and controls how far the observed data can be extrapolated. 
\par Another common example is the periodic covariance function, which is used to model a periodic function. It gives 
\[
K(t,t') = h^{2}\text{exp}\bigg(-\frac{2\text{sin}(\pi(t - t')/p)^{2}}{{l^{2}}}\bigg),
\]
where $p$ determines the distance between repetitions of a function, $h$ and $l$ are the same as in the squared exponential covariance function above. 
\par 
A wide variety of covariance functions has been proposed, see, for example, \citet{williams2006gaussian}, and this allows us great flexibilities in modelling among many different scenarios.  
\par With the assumption of the Gaussian distribution on a collection of all the discrete observations, $\boldsymbol{y} = [y(t_{1}),\ldots, y(t_{n})]^{T}$ follows an $n$-variate normal distribution, i.e.
\[
\boldsymbol{y} \sim  \mathcal{N}(\boldsymbol{\mu},\boldsymbol{K}),
\]
where $\boldsymbol{\mu} = [\mu(t_{1}),\ldots, \mu(t_{n})]^{T}$, and $\boldsymbol{K}$ is the $n \times n$ covariance matrix with entries $K(t_{i}, t_{j}) + \sigma^{2}\delta_{t_{i}, t_{j}}$. Here  $\delta_{t_{i}, t_{j}} = 1$ if $t_{i} = t_{j}$ and $0$ otherwise.

\par The log-likelihood function of the $n$-dimensional collection of the discrete observations $\boldsymbol{y}$ for all the (hyper-) parameters in the specified mean function and the specified covariance function (denoted by a generic ${\boldsymbol{\theta}}$) and the noise parameter $\sigma^{2}$ is
\begin{equation}\label{ch2eq: Gaussian log likelihood function}
	\mathcal{L}({\boldsymbol{\theta}},\sigma^{2}) = -\frac{1}{2}\text{log}|\boldsymbol{K}| -\frac{1}{2}(\boldsymbol{y} - \boldsymbol{\mu})^{T}(\boldsymbol{K})^{-1}(\boldsymbol{y} - \boldsymbol{\mu}) -\frac{n}{2} \text{log}(2\pi),
\end{equation}
where $|\cdot|$ denotes the determinant of a matrix.
\par Standard gradient-based numerical optimisation techniques, such as Conjugate Gradient method, can be used to maximise the log-likelihood function $\mathcal{L}({\boldsymbol{\theta}},\sigma^{2})$ in Equation (\ref{ch2eq: Gaussian log likelihood function}) to obtain the estimates of the model parameters.
\par For the corresponding value $y(t^{*})$ at any measurement time point $t^{*}$, we can denote the joint distribution of $[y(t_{1}),\ldots, y(t_{n}),y(t^{*})]^{T}$ as an $(n+1)$-variate normal distribution with a mean vector $[\mu(t_{1}),\ldots, \mu(t_{n}), \mu(t^{*}) ]^{T}$ and a covariance matrix as
\[
\begin{bmatrix}
	\boldsymbol{K} & \boldsymbol{K}^{*} \\
	\boldsymbol{K}^{*T} & K(t^{*},t^{*})+\sigma^{2}
\end{bmatrix},
\]
where $\boldsymbol{K}^{*} = [K(t^{*},t_{1}), \ldots,K(t^{*},t_{n})]^{T}$.
\par The conditional distribution of $y(t^{*})$ given $\boldsymbol{y}$ with the estimated (hyper-) parameters $\hat{\boldsymbol{\theta}}$ and noise variance $\hat{\sigma}^{2}$ through the optimisation of the log-likelihood function in Equation (\ref{ch2eq: Gaussian log likelihood function}), is then $\mathcal{N}(\hat{y}(t^{*}), \hat{\sigma}^{*2})$, where
\[
\hat{y}(t^{*}) = \mu(t^{*}) + \boldsymbol{K}^{*T} \boldsymbol{K}^{-1}(\boldsymbol{y} - \boldsymbol{\mu}),
\]
\[
\hat{\sigma}^{*2} = K(t^{*},t^{*})+ \hat{\sigma}^{2} - \boldsymbol{K}^{*T} \boldsymbol{K}^{-1} \boldsymbol{K}^{*}.
\]
\par The Gaussian process regression approach for conditioning of an unknown value by some realisations of a stochastic process in a joint Gaussian distribution can be seen as expanding discrete data from a function space point of view over the same domain. This idea can also be applied to the task of forecasting. With the Gaussian probabilistic framework, it can quantify the uncertainty associated with the historical experience of the observed data and then generate the full stochastic trajectories for out-of-sample forecasts with prediction intervals. We will discuss this in detail with applications to mortality and fertility data in the next section.
\section{Methodology} \label{sec: GPR Methodology}
\subsection{Gaussian process regression (GPR) model for mortality and fertility modelling and forecasting}
In this section, we introduce the proposed Gaussian process regression (GPR) model for mortality and fertility modelling and forecasting.
\par 
Let our discrete observed log mortality (or fertility) rates dataset be $\{t_{i}, y_{x}(t_{i})\}$ for $i = 1,\ldots,n$ and $x = 0,\ldots,m$, where $n$ is the total number of observed calendar years $t$, $m$ is the maximum age of interest, and $y_{x}(t_{i})$ is the log of mortality (or fertility) rates for a given age $x$ in a calendar year $t_{i}$. The proposed GPR model for mortality (or fertility) rates modelling and forecasting in a given age $x$ is 
\[
y_{x}(t_{i}) = f_{x}(t_{i}) + \epsilon_{x}(t_{i}), \hspace{0.5cm} {\epsilon_{x}(t_{i})} \sim \mathcal{N}(0, \sigma_{x}^{2}),
\]
where $f_{x}(t_{i})$ is the underlying function that needs to be estimated, $\{\epsilon_{x}(t_{i})\}_{i=1}^{n}$ allow the observation errors varying based on the assumption of i.i.d. normally distributed random variables with mean 0 and constant variance $\sigma_{x}^{2}$ for a given age $x$.
\par Following the Gaussian process regression paradigm discussed in Section \ref{sec: GPR Theoretical background of Gaussian process regression}, the underlying function $\{f_{x}(t) : t \in\mathcal{T} \}$ for mortality (or fertility) modelling and forecasting of each  age $x$ is assumed to follow a Gaussian process with a specified prior mean function $\mu_{x}(t)$ and a specified prior covariance function $K_{x}(t,t')$ over the domain $\mathcal{T}$\footnote{Note that mortality and fertility data are not directly of a functional nature, here we assume that there are underlying functional time series which are observed with errors at discrete points.}. It gives for all $t \in \mathcal{T}$
\begin{equation}{\label{eq: Proposed GPR model}}
	f_{x}(t) \sim \mathcal{GP}(\mu_{x}(t),K_{x}(t,t')).
\end{equation}
\subsubsection{Specified Gaussian process prior mean function}
Under the Gaussian process regression paradigm, we are allowed to choose a prior mean function that reflects our prominent belief in the unknown function $f_{x}(t)$ for the Gaussian process regression model. Due to incomplete information on the functional form among time-series age-specific mortality and fertility rates over the domain $\mathcal{T}$, and a wide range of patterns in the age-specific mortality and fertility rates exhibiting in our numerical examples in Section \ref{sec: Applications} (see Figure {\ref{fig: Univariate time series of log male mortality rates with 20-year age intervals}} and Figure \ref{fig: Univariate time series of log fertility rates with 5-year age intervals}), we specify and adopt a natural cubic spline function as the non-parametric mean function in the proposed GPR model for mortality and fertility modelling and forecasting. The justifications for this choice here are that the natural cubic spline function can handle a broad range of complex and non-linear functions that exist in all the age-specific mortality and fertility rates as a generalised and unified approach. It behaves like a higher-order polynomial regression model for curve fitting with no presumption needed on the relationship between the covariate and response variables. Meanwhile, it can extrapolate the trend of historical data smoothly into the future in an appropriate direction without suffering the overfitting problem that occurs commonly in the higher-order polynomial regression model when it comes to forecasting. 
\par To explain further how it works mathematically, we consider the natural cubic spline function with strictly increasing $K$ knots $\lbrace\xi_{k}\rbrace_{k = 1}^{K}$ where $\xi_{1}< \xi_{2}< \ldots < \xi_{K}$ over the observed time interval $[t_{1},t_{n}]$ can be represented as the truncated power series representation for a cubic spline function. It gives
\begin{equation}\label{eq: Truncated power series representation for a cubic spline function}
	\mu_{x}(t)=\sum_{p=0}^3\alpha_{p}t^{p} + \sum_{k=1}^K\beta_{k}(t - \xi_{k})^{3}_{+},
\end{equation}
with additional constraints on the boundary conditions when $t < \xi_{1}$ and $t \geq \xi_{K}$, such that
\begin{equation}\label{eq: Additional constraints}
	\alpha_{2} = 0, \alpha_{3} = 0,  \\
	\sum_{k=1}^K\beta_{k} = 0, \text{and} \sum_{k=1}^K\beta_{k}\xi_{k} = 0,
\end{equation}
where $(\cdot)_{+}$ denotes the positive part, $\{{\alpha_{p}}\}_{p=0}^{3}$ are the coefficients of the cubic polynomial and $\{\beta_{k}\}_{k=0}^{K}$ are the coefficients of the truncated power series for the cubic splines with $K$ interior knots.
\par  By imposing the above boundary constraints in Equation (\ref{eq: Additional constraints}), the natural cubic spline function can be further derived as a reduced basis when $t \geq \xi_{K}$ for extrapolation as
\begin{equation}\label{eq: natural cubic spline function for ex}
	\begin{aligned}[b]
		\mu_{x}(t) &=\sum_{p=0}^3\alpha_{p}t^{p} +\sum_{k=1}^K\beta_{k}(t - \xi_{k})^{3}\\
		&=\sum_{p=0}^3\alpha_{p}t^{p} + \sum_{k=1}^{K}\beta_{k}(t^{3} - 3\xi_{k}t^{2} + 3{\xi^{2}_{k}t - \xi^{3}_{k}})\\
		&= \alpha_{0}+\alpha_{1}t + \underbrace{\alpha_{2}t}_{=0} + \underbrace{\alpha_{3}t}_{=0} + \underbrace{\sum_{k=1}^{K}\beta_{k}t^{3}}_{=0} - \underbrace{\sum_{k=1}^{K}3\beta_{k}\xi_{k}t^{2}}_{=0} + \sum_{k=1}^{K}3\beta_{k}\xi_{k}^{2}t - \sum_{k=1}^{K}\beta_{k}\xi_{k}^{3}\\
		&= \alpha_{0}+\alpha_{1}t + \sum_{k=1}^{K}3\beta_{k}\xi_{k}^{2}t - \sum_{k=1}^{K}\beta_{k}\xi_{k}^{3}\\
		&= (\alpha_{0}-\sum_{k=1}^{K}\beta_{k}\xi_{k}^{3} ) + (\alpha_{1} + \sum_{k=1}^{K}3\beta_{k}\xi_{k}^{2})t.
	\end{aligned}
\end{equation}
For simplicity of notation, we denote  $ c_{0} = (\alpha_{0}-\sum_{k=1}^{K}\beta_{k}\xi_{k}^{3})$ and $c_{1} =  (\alpha_{1} + \sum_{k=1}^{K}3\beta_{k}\xi_{k}^{2})$. Then
\[
\mu_{x}(t) = c_{0} + c_{1}t,
\]
thus $\mu_{x}(t)$ is a linear function with constants $c_{0}$ and $c_{1}$ when $t \geq \xi_{K}$ for extrapolation. 
\par Unlike the cubic spline function whose behaviour tends to be erratic near boundaries and the extrapolation can be unrealistic outside the observed time interval $[t_{1},t_{n}]$, the natural cubic spline function forces its extrapolation outside regions beyond the observed data to be a linear function with zero second derivative that agrees with the spline in its intercept and slope values among its last knot \citep{friedman2001elements}. With this distinctive feature, the natural cubic spline function can hence inherit the local information of recent data and gives a reasonable direction for extrapolation. In demographic modelling, it is often the case that more recent experience has greater relevance on future behaviour than those data from the distant past. On account of this, we specify the natural cubic spline function as the mean function for mortality and fertility modelling and forecasting in the proposed GPR model. 
\subsubsection{Specified Gaussian process prior covariance function}
In the proposed GPR model, the specified covariance function is used to discover and capture the similarities and various time-correlated structures among the historical demographic trends for different age groups, then project their patterns into the future.
\par Given that the choice of covariance functions in Gaussian process regression remains an ongoing research problem, the choice of a covariance function is based on the empirical evidence by simulating the GPR model with different covariance functions in our study. Apart from the squared exponential and the periodic covariance functions mentioned in Section \ref{sec: GPR Theoretical background of Gaussian process regression}, we have also considered and tested four more common covariance functions, including the rational quadratic, Mat\'{e}rn 3/2 and Mat\'{e}rn 5/2 covariance functions documented in the Gaussian process literature \citep{williams2006gaussian}, and the spectral mixture covariance function proposed by \citet{wilson2013gaussian}. The spectral mixture covariance function is derived by modelling a spectral density of the Fourier transform of input-correlated structures with a Gaussian mixture, which can automatically discover patterns and extrapolate far beyond the available data; see \citet{wilson2013gaussian} for detailed discussions. 
\par From our empirical experiments\footnote{The empirical experiments are discussed in Section \ref{sec: Applications}. As the differences in the root mean square error (RMSE) of using different covariance functions are insignificant with around $\pm 0.005$, we do not present the test results.}, we have found that the spectral mixture covariance function is indeed relatively more successful in capturing different patterns of age-specific mortality and fertility rates. It also provides more accurate forecast results than the other covariance functions for mortality and fertility predictions. \citet{wu2018gaussian} also discovered the similar results. The spectral mixture covariance function is thus selected as the primary covariance function in the proposed GPR model for all age-specific mortality and fertility rates as a unified approach. \par Considering a $Q$ number of Gaussian mixture components, in which the $q$-th component has mean $\lambda_{q}$ and variance $\nu_{q}^{2}$, the spectral mixture covariance function is defined as
\begin{equation}\label{eq: SM covariance function}
	K_{x} (t,t') = \sum_{q=1}^{Q}w_{q}\text{exp}\{-2\pi^{2}(t - t')^{2}\nu_{q}^{2}\}\text{cos}(2\pi(t - t')\lambda_{q}),
\end{equation}
where $w_{q}$ is the weight specifying the contribution of the $q$-{th} Gaussian mixture component.

\subsubsection{Likelihood function of the proposed GPR model}
It yields that $\boldsymbol{y}_{x} = [y_{x}(t_{1}),\ldots, y_{x}(t_{n})]^{T}$ in each given age $x$ follows an $n$-variate normal distribution, such that
\[
\boldsymbol{y}_{x} \sim  \mathcal{N}(\boldsymbol{\mu}_{x},\boldsymbol{K}_{x}),
\]
where $\boldsymbol{\mu}_{x} = [\mu_{x}(t_{1}),\ldots, \mu_{x} (t_{n})]^{T}$ and $\boldsymbol{K}_{x} $ is the $n \times n$ specified covariance matrix with entries $K_{x} (t_{i},t_{j}) + \sigma_{x}^{2}\delta_{t_{i},t_{j}}$. Here  $\delta_{t_{i},t_{j}} = 1$ if $t_{i}=t_{j}$ and $0$ otherwise. 
\par The log-likelihood function of the $n$-dimensional collection of the discrete observations $\boldsymbol{y}_{x}$ in each age $x$ for all the (hyper-) parameters (denoted by a generic ${\boldsymbol{\theta}}_{x}$) and the noise variance $\sigma_{x}^{2}$ is
\begin{equation}\label{eq: Gaussian process log likelihood function}
	\mathcal{L}({\boldsymbol{\theta}}_{x},\sigma_{x}^{2}) = -\frac{1}{2}\text{log}|\boldsymbol{K}_{x}| -\frac{1}{2}(\boldsymbol{y}_{x} - \boldsymbol{\mu}_{x})^{T}(\boldsymbol{K}_{x})^{-1}(\boldsymbol{y}_{x} - \boldsymbol{\mu}_{x}) -\frac{n}{2} \text{log}(2\pi),
\end{equation}
where $|\cdot|$ is the determinant of a matrix. \par 
Standard gradient-based numerical optimisation techniques can be used to maximise the log-likelihood function $\mathcal{L}({\boldsymbol{\theta}}_{x},\sigma_{x}^{2})$ in Equation (\ref{eq: Gaussian process log likelihood function}) to obtain the estimates of the model parameters.
\subsubsection{Out-of-sample forecasts and prediction intervals of the proposed GPR model}
Let $t^{*}_{h}$ be the $h$-step ahead of the last observed calendar year where $t_{n} < t^{*}_{1} <\ldots< t^{*}_{h}$ for all $t_{i}^{*} \in \mathcal{T}$. Then the joint distribution of $[y_{x}(t_{1}),\ldots, y_{x}(t_{n}),y_{x}(t_{1}^{*}),\ldots,y_{x}(t_{h}^{*})]^{T}$ is an $(n+h)$-variate normal distribution that comes with the mean vector $[\mu_{x}(t_{1}),\ldots,\mu_{x}(t_{n}),$ $\mu_{x}(t_{1}^{*}),\ldots,\mu_{x}(t_{h}^{*})]^{T}$ and the covariance matrix is
\[
\begin{bmatrix}
	\boldsymbol{K}_{x} & \boldsymbol{K}_{x}^{*} \\
	\boldsymbol{K}_{x}^{*T} & \boldsymbol{K}_{x}^{**}
\end{bmatrix},
\]
where $\boldsymbol{K}_{x}^{*}$ is the size of $n \times h$ covariance matrix, and $\boldsymbol{K}_{x}^{**}$ is the size of $h \times h$ covariance matrix with entries $K_{x}(t_{i}^{*},t_{j}^{*}) + \sigma_{x}^{2}\delta_{t_{i}^{*},t_{j}^{*}}$. Here $\delta_{t_{i}^{*},t_{j}^{*}} = 1$ if $t_{i}^{*} = t_{j}^{*}$ and $0$ otherwise. 
\par Therefore, the $h$-step ahead out-of-sample forecast and its variance of the age-specific mortality (or fertility) rates can be found from the conditional distribution of $\boldsymbol{y}_{x}^{*} = [y_{x}(t_{1}^{*}),\ldots,y_{x}(t_{h}^{*})]^{T}$ given $\boldsymbol{y}_{x}$ with the estimated (hyper-) parameters $\hat{\boldsymbol{\theta}}_{x}$ and noise variance $\hat{\sigma}_{x}^{2}$ through the optimisation of the log-likelihood function in Equation (\ref{eq: Gaussian process log likelihood function}), is then $\mathcal{N}(\hat{\boldsymbol{y}}^{*}_{x}, \text{Var}({\boldsymbol{y}}^{*}_{x}))$, and let $\boldsymbol{\mu}_{x}^{*} = [\mu_{x}(t_{1}^{*}),\ldots,\mu_{x}(t_{h}^{*})]^{T}$, where
\begin{equation}\label{ch4eq: Conditional mean for Gaussian process regression}
	\hat{\boldsymbol{y}}^{*}_{x} = \boldsymbol{\mu}_{x}^{*} +  \boldsymbol{K}_{x}^{*T} \boldsymbol{K}_{x}^{-1}(\boldsymbol{y}_{x} - \boldsymbol{\mu}_{x}),
\end{equation}
\begin{equation}\label{ch4eq: Conditional variance for Gaussian process regression}
	\text{Var}({\boldsymbol{y}}^{*}_{x}) = \boldsymbol{K}_{x}^{**}  - \boldsymbol{K}_{x}^{*T} \boldsymbol{K}_{x}^{-1} \boldsymbol{K}^{*}_{x} + \hat{\sigma}_{x}^{2}\boldsymbol{I}.
\end{equation}

\par With the normality assumptions on the model error and the known $\text{Var}({\boldsymbol{y}}^{*}_{x})$, a $100(1-\alpha)\%$ prediction  interval for ${\boldsymbol{y}}^{*}_{x}$ can be calculated as ${\boldsymbol{y}}^{*}_{x}\pm z_{\alpha}\sqrt{\text{Var}({\boldsymbol{y}}^{*}_{x})}$, where $z_{\alpha}$ is the $(1-\alpha/2)$ quantile of the standard normal distribution. 

\section{Applications} \label{sec: Applications}
In this section, we demonstrate the abilities of the proposed GPR model for modelling and forecasting two different sets of demographic data  $-$ mortality data and fertility data. We first apply the proposed model to the male mortality data and the fertility data of Japan for illustration purposes. We then compare and evaluate the quality of the fitted mortality and fertility curves by the proposed method with some other existing approaches using the mortality and fertility data of ten different developed countries.
\subsection{Male mortality data }
The male mortality data of Japan are available from the year 1947 to the year 2016 from the \citet{human2017university}. The database consists of the age-specific male mortality rates by a single calendar year of age up to 110 years old. The age-specific male mortality rates are defined as the number of deaths in males during a calendar year, proportional to the male resident population of the same age during the same calendar year. We restrict our experimental mortality data up to age 100 to avoid any potential problem associated with the erratic mortality rates above age 100.
\par
\begin{figure}[!thb]
	\centering
	\begin{minipage}{1.05\textwidth}
		\centering
		\includegraphics[width=1\linewidth]{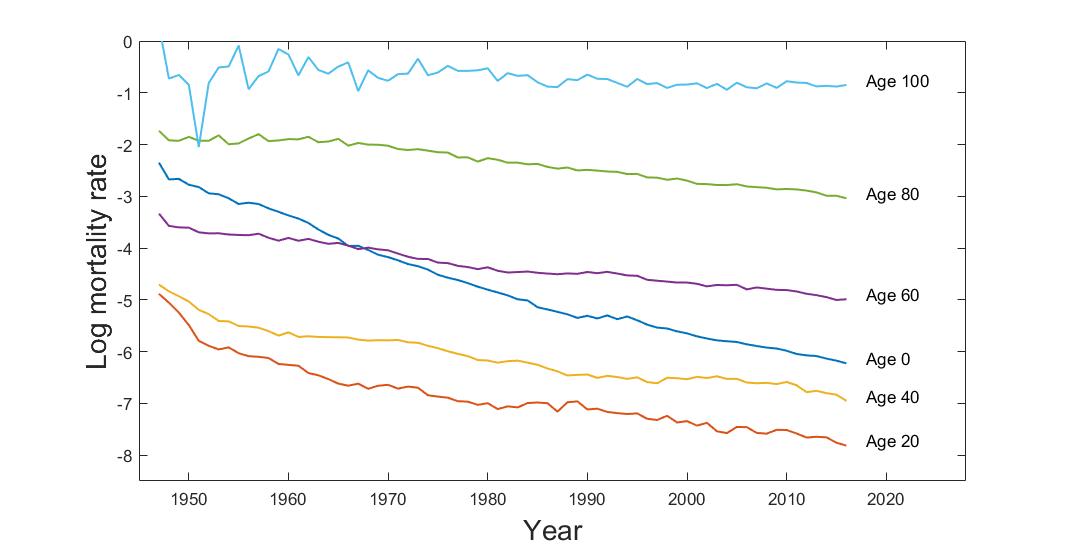}
		\subcaption*{(a)}
		\phantomcaption
		\label{fig: Univariate time series of log male mortality rates with 20-year age intervals}
	\end{minipage}
	\\
	\begin{minipage}{1.05\textwidth}
		\centering
		\includegraphics[width=1\linewidth]{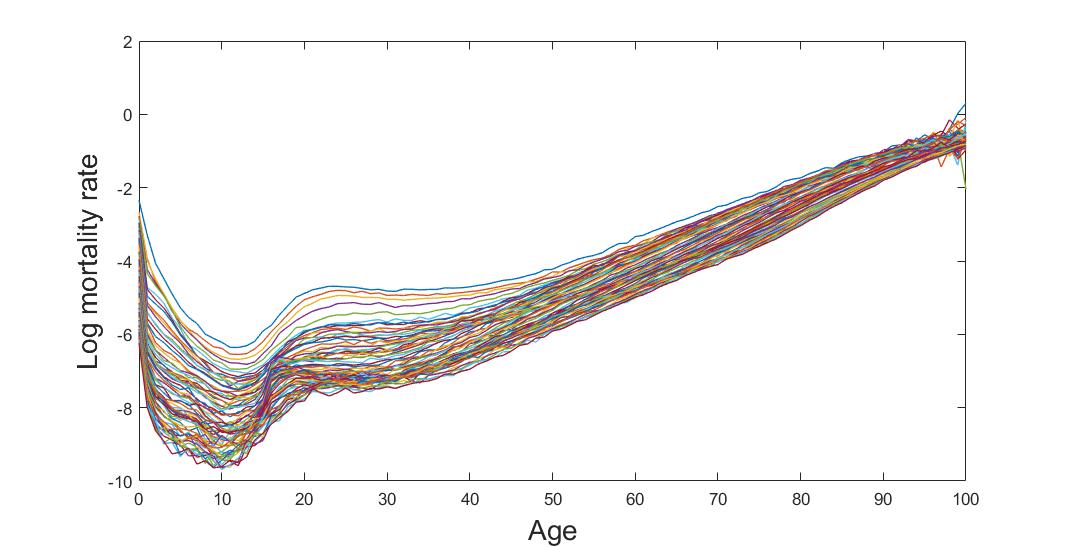}
		\subcaption*{(b)}
		\phantomcaption
		\label{fig: Log male mortality curves}
	\end{minipage}
\caption{(a) Univariate time series of the log age-specific male mortality rates with 20-year age intervals and (b) the log male mortality curves from age 0 to age 100 from the year 1947 to the year 2016 in Japan.}
\end{figure}

 The observed male mortality data are presented in Figure \ref{fig: Univariate time series of log male mortality rates with 20-year age intervals} as separate univariate time series of the log age-specific male mortality rates with 20-year age intervals from age 0 to age 100 from the year 1947 to the year 2016. We can see that there is a general decrease in male mortality rates among all the selected age groups during the examined period. The decline in mortality rates at higher ages seems to change more slowly than those at younger ages. Figure \ref{fig: Log male mortality curves} presents the log male mortality curves, which give us information about the general trends and the variations of the age-specific male mortality rates over the observed period. There is an apparent hump around 18 to 25 years old which usually relates to reckless behaviour in teenage ages and a general drop for all population in mortality rates over time, due primarily to the advances in medical technology.
\subsection{Mortality modelling and forecasting}
\par In the demonstration of the proposed model, we aim to display 10-years-ahead out-of-sample forecasts of mortality rates. We first split the dataset into a training dataset with the observed mortality rates from the year 1947 to the year 2006 and a test dataset with the remaining observed mortality data from the year 2007 to the last sample data in the year 2016. We use our GPR model in Equation (\ref{eq: Proposed GPR model}) to fit the male mortality rates for each given age in the training dataset. We equally place the interior knots $\xi$ inside the quantile position of the training time interval $(K = 4)$ in the natural cubic spline mean function in Equation (\ref{eq: Truncated power series representation for a cubic spline function}), and use two Gaussian mixture components $(Q = 2)$ in the spectral mixture covariance function in Equation (\ref{eq: SM covariance function}). These settings are the minimal requirements for the number of parameters needed for the Gaussian process mean function and covariance function. Further tests on the numbers of knots $K$ and Gaussian mixture components $Q$ have been made, and no significant difference in our experiment results was found when $K > 4$ and $Q > 2$. The parameters $\boldsymbol{\theta}_{x}$ including the parameters of the mean function $\boldsymbol{\theta}_{\mu_{x}}$ and the hyperparameters of the covariance function $\boldsymbol{\theta}_{K_{x}} = \lbrace(w_{q}, m_{q}, \nu_{q})\rbrace_{q=1}^{Q}$ and the noise variance ${\sigma}_{x}^{2}$ can be estimated by maximising the log likelihood function in Equation (\ref{eq: Gaussian process log likelihood function}) for each given age $x$. However, estimating all unknown parameters involved both in the mean and the covariance structures for all age groups can be time demanding with high computational costs in practice. Following the suggestions of \citet{shi2007gaussian}, we remedy this problem by computing the parameters of the mean function through the ordinary least squares approach, then estimating the parameters of the covariance function by maximising the log likelihood function in Equation (\ref{eq: Gaussian process log likelihood function}). This estimating procedure is analogous to de-trending the age-specific mortality rates by the empirical mean function then modelling the residuals by a zero-mean Gaussian process. It indeed improved the stability and the speed of the estimation procedure for all (hyper-) parameters in our numerical examples. \par
\begin{figure}[!thb]
	\begin{minipage}{0.5\textwidth}
		\centering
		\includegraphics[width=1\linewidth]{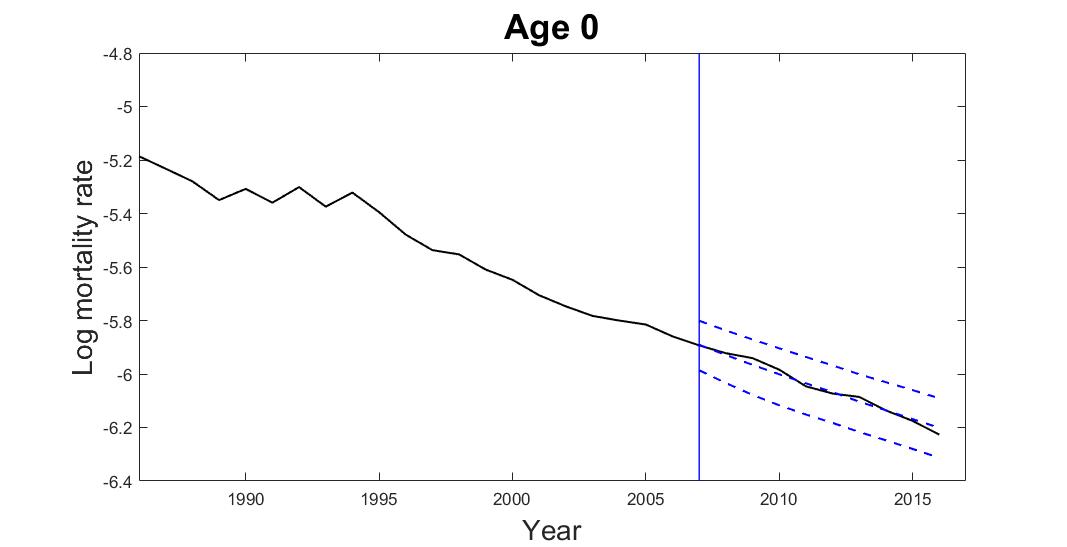}
	\end{minipage}
	\begin{minipage}{0.5\textwidth}
		\centering
		\includegraphics[width=1\linewidth]{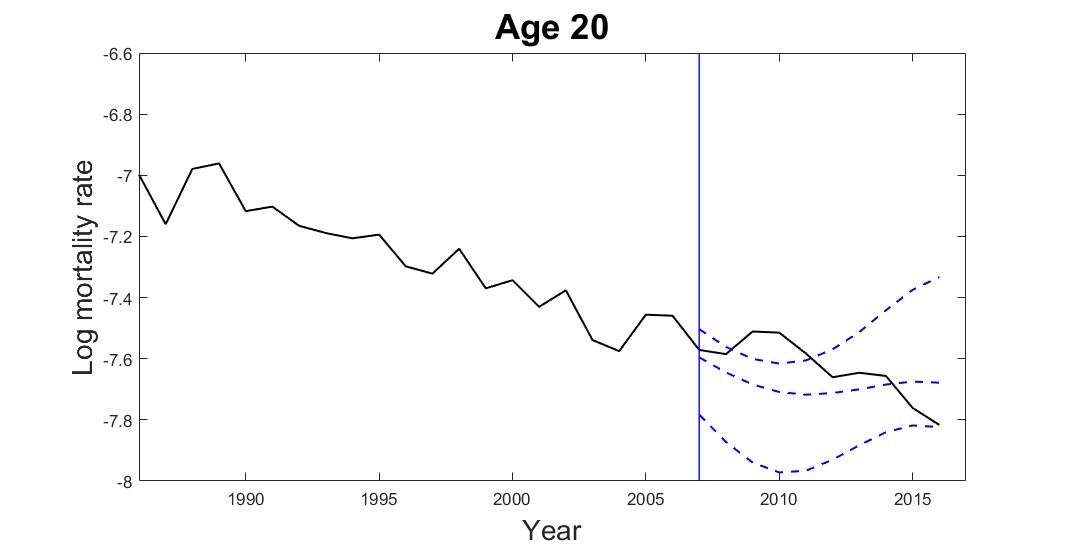}
	\end{minipage} 
	\begin{minipage}{0.5\textwidth}
		\centering
		\includegraphics[width=1\linewidth]{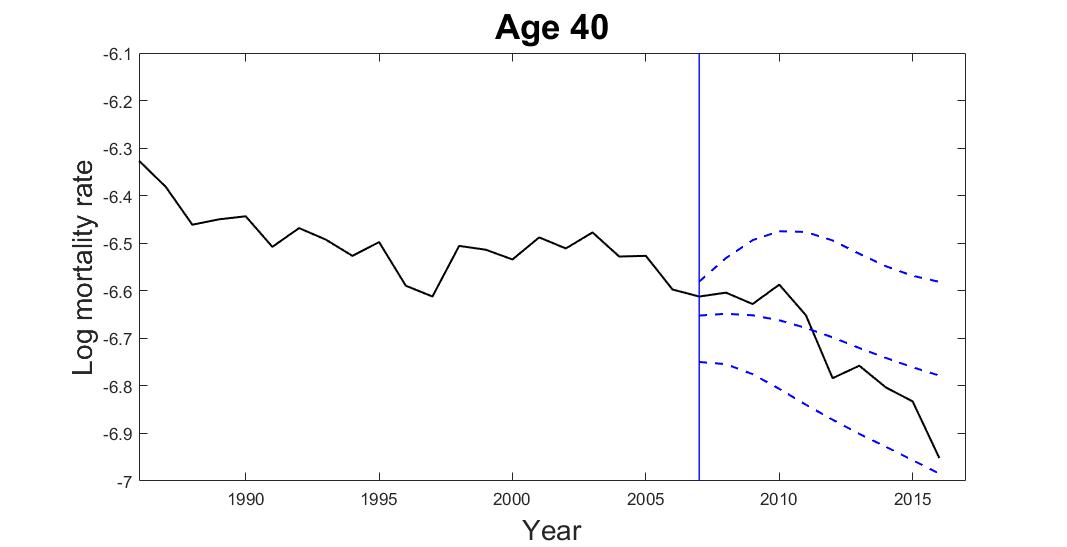}
	\end{minipage}
	\begin{minipage}{0.5\textwidth}
		\centering
		\includegraphics[width=1\linewidth]{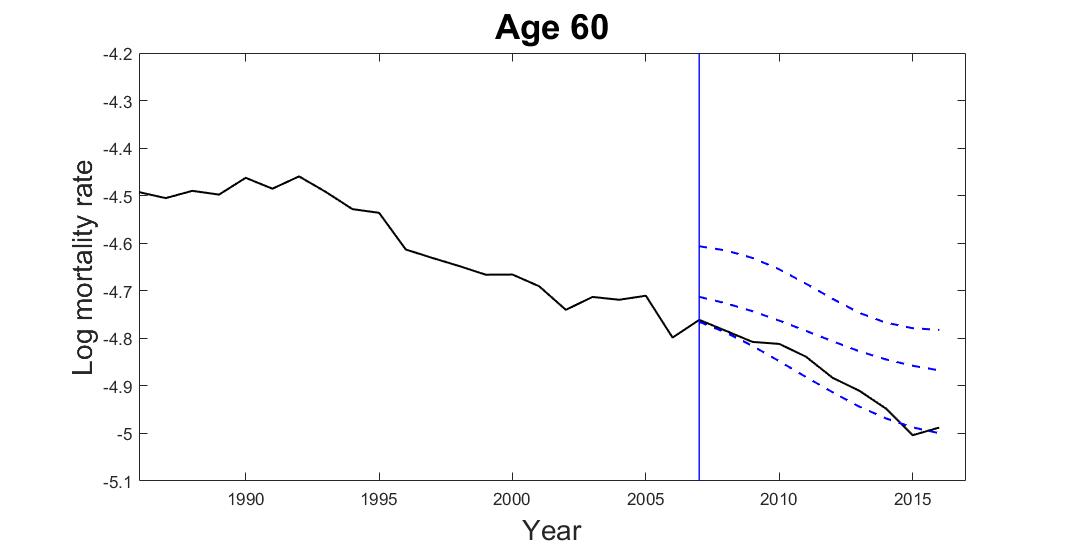}
	\end{minipage} 
	\begin{minipage}{0.5\textwidth}
		\centering
		\includegraphics[width=1\linewidth]{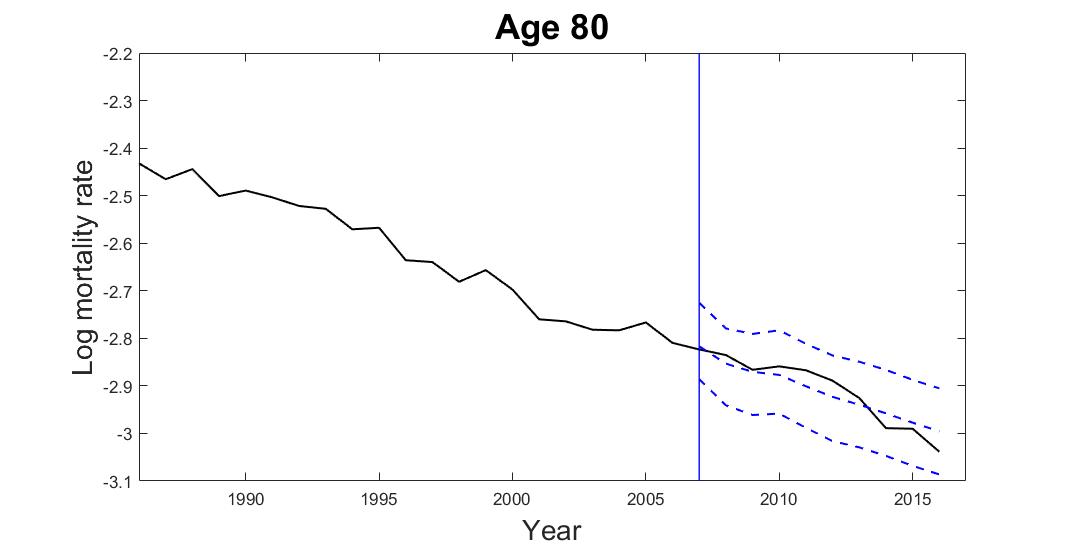}
	\end{minipage}
	\begin{minipage}{0.5\textwidth}
		\centering
		\includegraphics[width=1\linewidth]{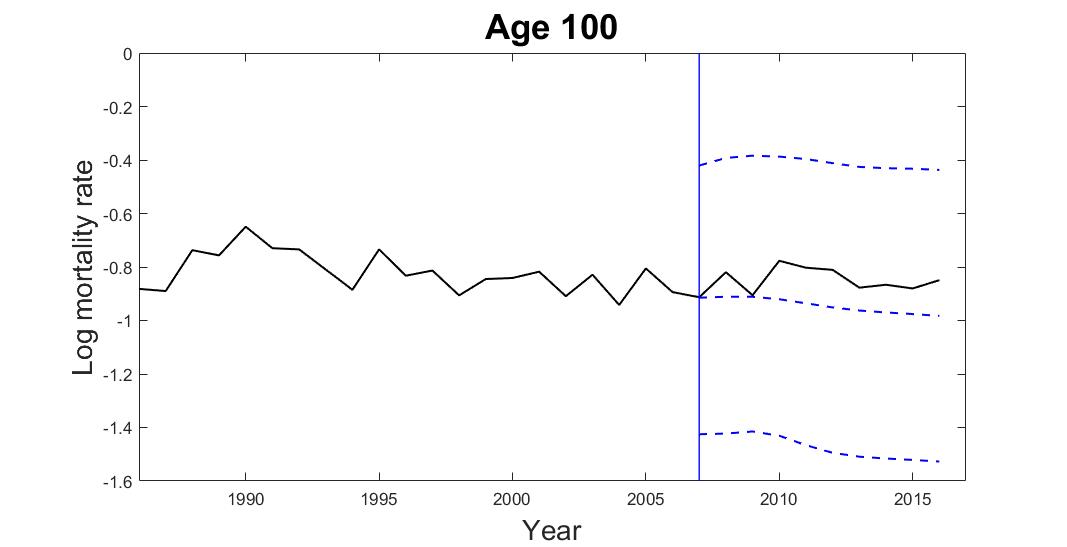}
	\end{minipage} 
	\caption{Predicted male mortality rates of the selected age groups from age 0 to age 100 with 20-year age intervals using the proposed GPR model for the year 2007 to the year 2016 based on the observations from the year 1947 to the year 2006 in Japan. The solid lines are the observed values, and the dashed lines are the predictive mean and the 95\% prediction intervals. The vertical line indicates the starting point of the predictions.} 
	\label{fig: Predicted male mortality rates of selected age groups from age 0 to age 100 with 20-year age intervals using the proposed GPR model}
\end{figure}

\begin{figure}[!thb]
	\centering
	\includegraphics[width=1\linewidth]{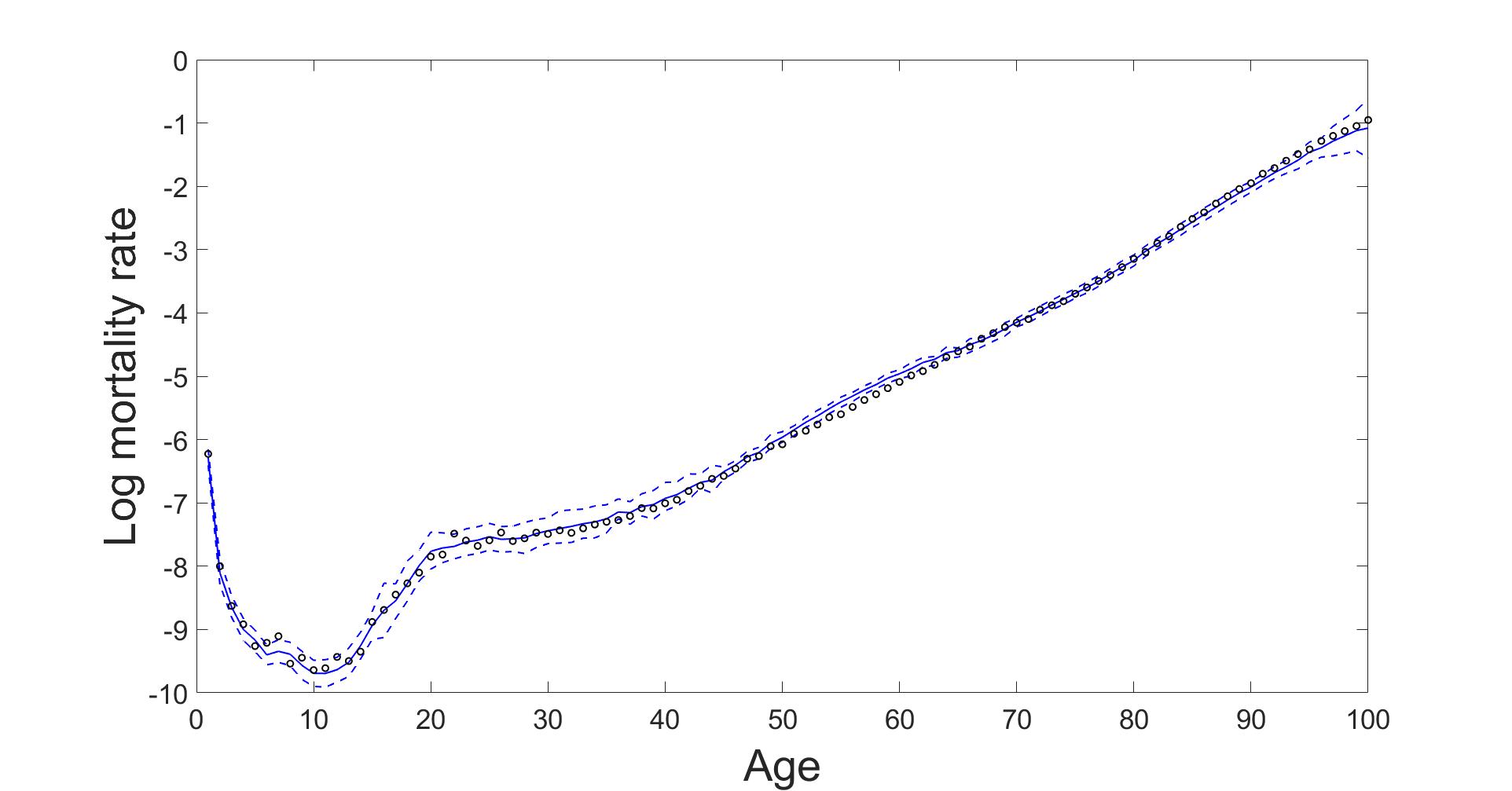}
	\caption{Predicted male mortality curve from age 0 to age 100 with 95\% prediction intervals using the proposed GPR model for the year 2016 based on the observations from the year 1947 to the year 2006 in Japan. The circles are the true log mortality rates, the solid line is the predictive mean, and the dashed lines are the 95\% prediction intervals.}
	\label{fig: Predicted male mortality curve} 
\end{figure}

 The predictive mean and variance for forecasting age-specific mortality rates can be obtained by Equation (\ref{ch4eq: Conditional mean for Gaussian process regression}) and Equation (\ref{ch4eq: Conditional variance for Gaussian process regression}) respectively. Figure \ref{fig: Predicted male mortality rates of selected age groups from age 0 to age 100 with 20-year age intervals using the proposed GPR model} demonstrates the predicted results of the selected age-specific mortality rates by the proposed GPR model. We can see that the proposed model can capture the varying patterns in mortality rates among different age groups properly. To construct a predicted mortality curve for any specified year, we extract the predicted age-specific mortality rates from age 0 to age 100 in that specified year. Figure \ref{fig: Predicted male mortality curve} gives an example of the 10-years-ahead out-of-sample forecast results of the male mortality curve with the 95\% prediction intervals using the proposed GPR model for the year 2016 based on the observations from the year 1947 to the year 2006.
 To construct a predicted mortality curve for any specified year, we extract the predicted age-specific mortality rates from age 0 to age 100 in that specified year. Figure \ref{fig: Predicted male mortality curve} gives an example of the 10-years-ahead out-of-sample forecast results of the male mortality curve with 95\% prediction intervals using the proposed GPR model for the year 2016 based on the observations from the year 1947 to the year 2006. 
\subsection{Fertility data}
We move to the second case study of modelling and forecasting the age-specific fertility rates. The fertility data of Japan are available from the year 1947 to the year 2016 from the \citet{humanfertility2017university}. The database consists of the age-specific fertility rates by calendar year from 12 to 55 years old. The age-specific fertility rates are defined as the number of births during a calendar year, based on the age of the mother proportional to the total number of the female resident population. Our study focuses on the age-specific fertility rates from age 15 to age 45, given that this age range is closer to reality \citep{hyndman2007robust}, and the problem of missing data is also common when the age-specific fertility data starts from age 46 to age 55. Excluding this age range not just helps stabilise the fitted fertility curves for illustration in this section, it also gives fairer forecast performance comparisons of the proposed model to other existing models without making any subjective adjustments on the missing data when involving fertility data of several other countries which will be discussed in Section \ref{sec: Comparisons and forecast accuracy evaluations with existing models}. 
\par
\begin{figure}[!thb]
	\centering
	\begin{minipage}{1.05\textwidth}
		\centering
		\includegraphics[width=1\linewidth]{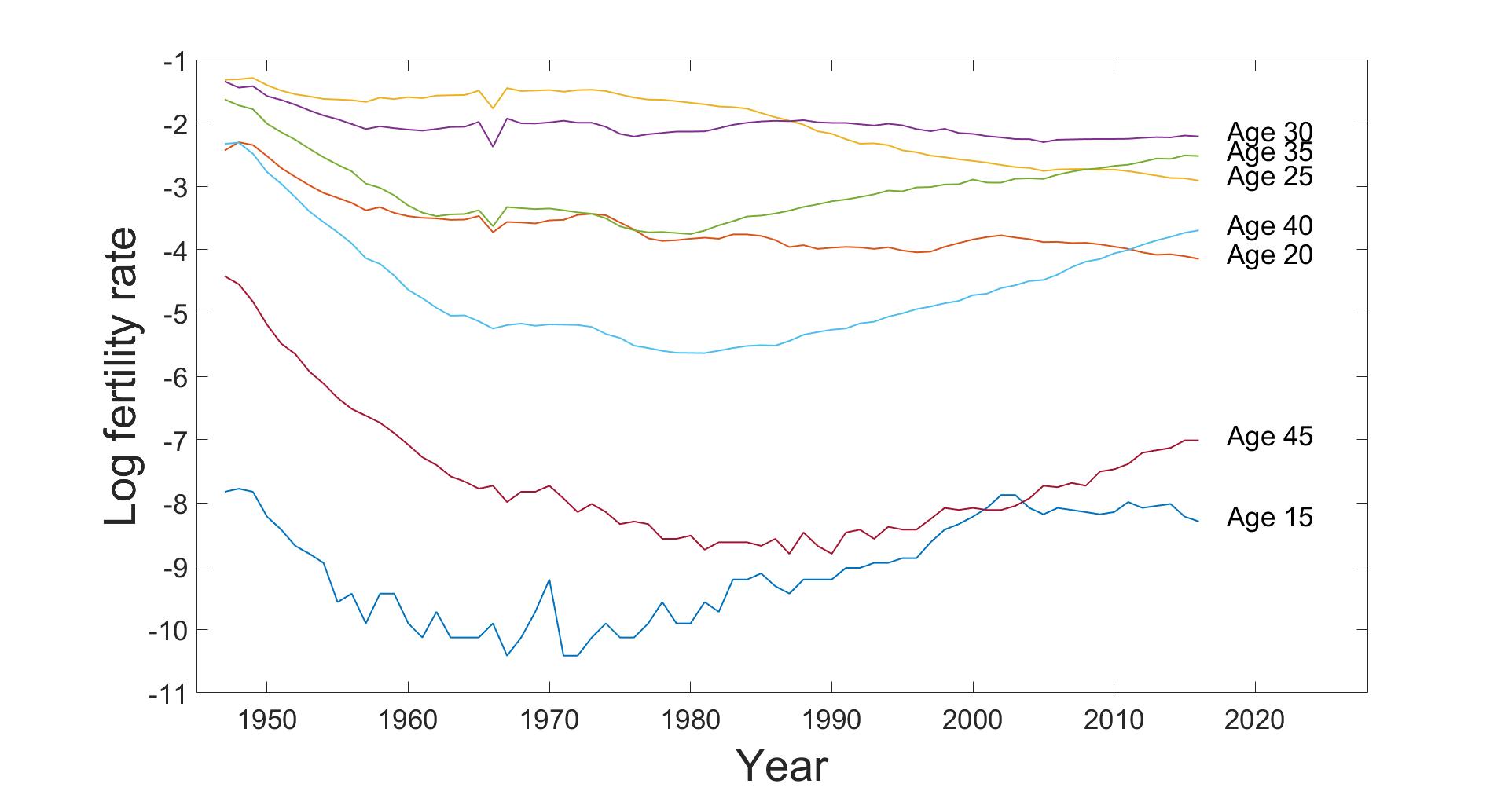}
		\subcaption*{(a)}
		\phantomcaption
		\label{fig: Univariate time series of log fertility rates with 5-year age intervals}
	\end{minipage}
	
	\begin{minipage}{1.05\textwidth}
		\centering
		\includegraphics[width=1\linewidth]{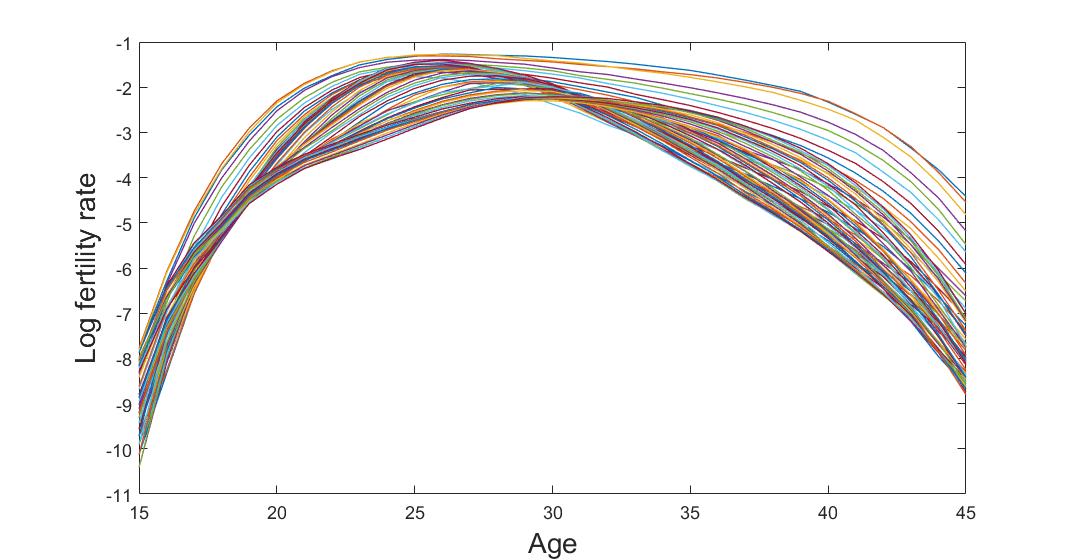}
		\subcaption*{(b)}
		\phantomcaption
		\label{fig: Log fertility curve from age 15 to age 45 from 1947 to 2016 of Japan.}
	\end{minipage}
\caption{ (a) Univariate time series of the log fertility rates with 5-year age intervals and (b) the log fertility curves from age 15 to age 45 from the year 1947 to the year 2016 in Japan.}
\end{figure}

 We present the historical fertility data of Japan as separate univariate time series of the log age-specific fertility rates with 5-year age intervals in Figure \ref{fig: Univariate time series of log fertility rates with 5-year age intervals} and as the log fertility curve from age 15 to age 45 from the year 1947 to the year 2016 in Figure \ref{fig: Log fertility curve from age 15 to age 45 from 1947 to 2016 of Japan.}. They reflect the patterns of fertility change caused by social conditions in different periods in Japan. For example, there was a `baby boom' in all age groups after the end of World War II in 1945 and a rapid decrease in the birth rate during the `Japanese economic miracle period' in the early 1980s due to a delay in child-bearing while the economy was establishing rapidly at that time. In more recent years, there was an increasing trend in fertility at higher ages, because the females devoted more time in their educations and careers. In general, females from age 20 to age 40 have relatively higher fertility rates compared to the age groups 15 and 45. The bunch of fertility curves in Figure \ref{fig: Log fertility curve from age 15 to age 45 from 1947 to 2016 of Japan.} display a concave shape. It shows that the fertility rates climb from age 15 and reach their peak at around age 30 then decline. There exist some sparse patterns in the later part of the fertility curves, which reflects that the variations in birth rates above age 35 are obvious across the observed years.
\subsection{Fertility modelling and forecasting}
\par
\begin{figure}[!thb]
	\begin{minipage}{0.5\textwidth}
		\centering
		\includegraphics[width=1\linewidth]{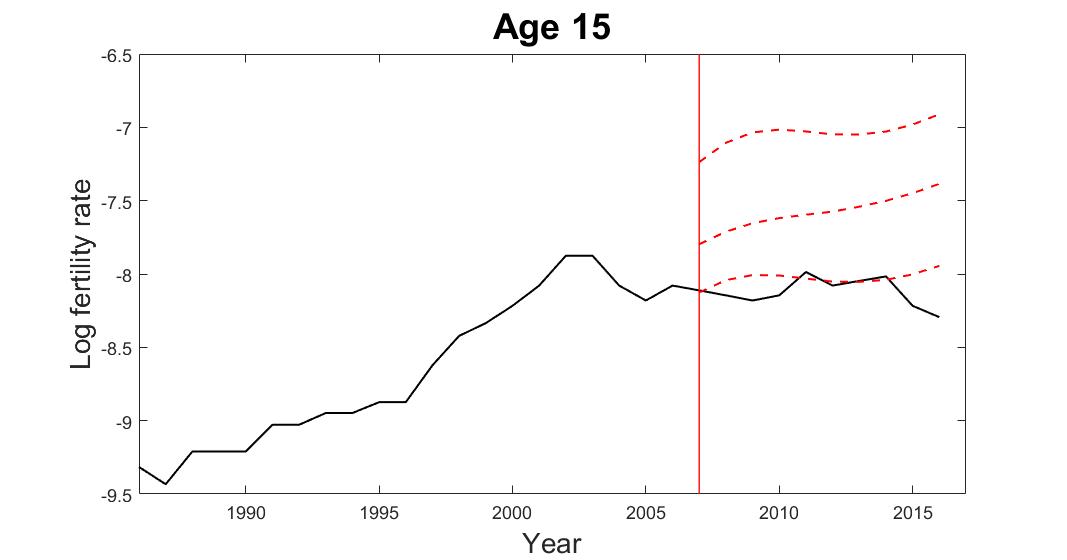}
	\end{minipage}
	\begin{minipage}{0.5\textwidth}
		\centering
		\includegraphics[width=1\linewidth]{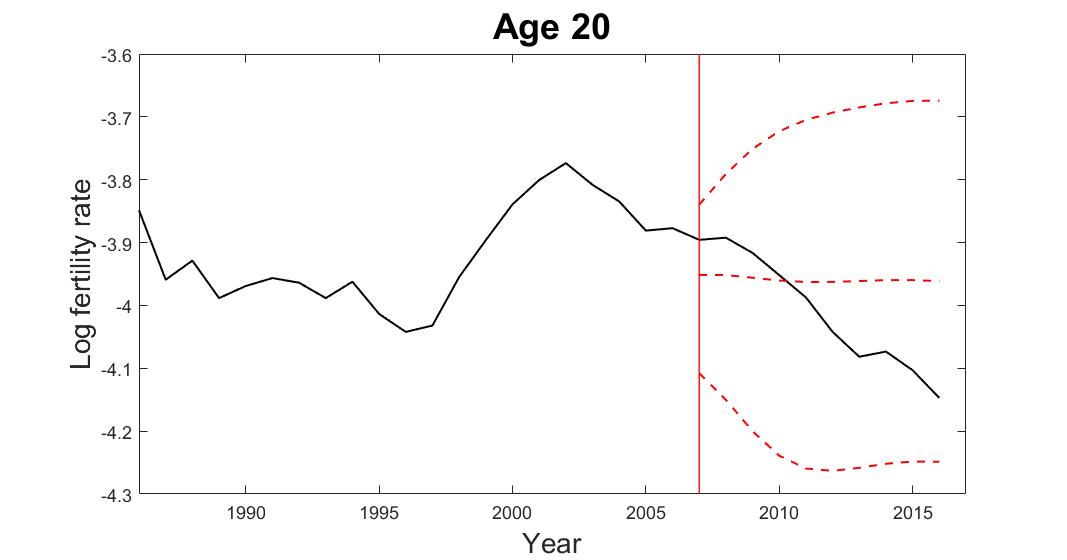}
	\end{minipage} 
	\begin{minipage}{0.5\textwidth}
		\centering
		\includegraphics[width=1\linewidth]{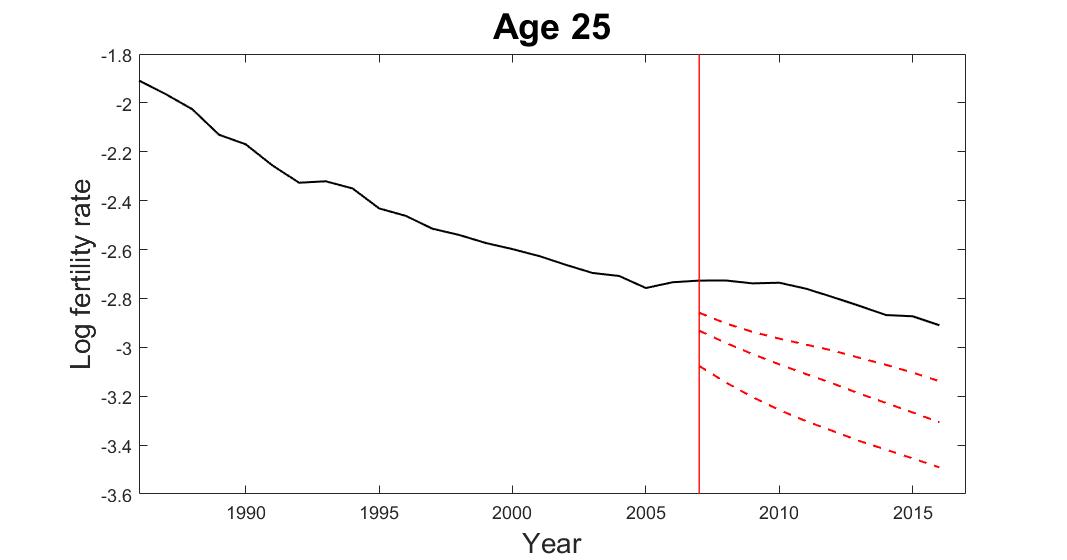}
	\end{minipage}
	\begin{minipage}{0.5\textwidth}
		\centering
		\includegraphics[width=1\linewidth]{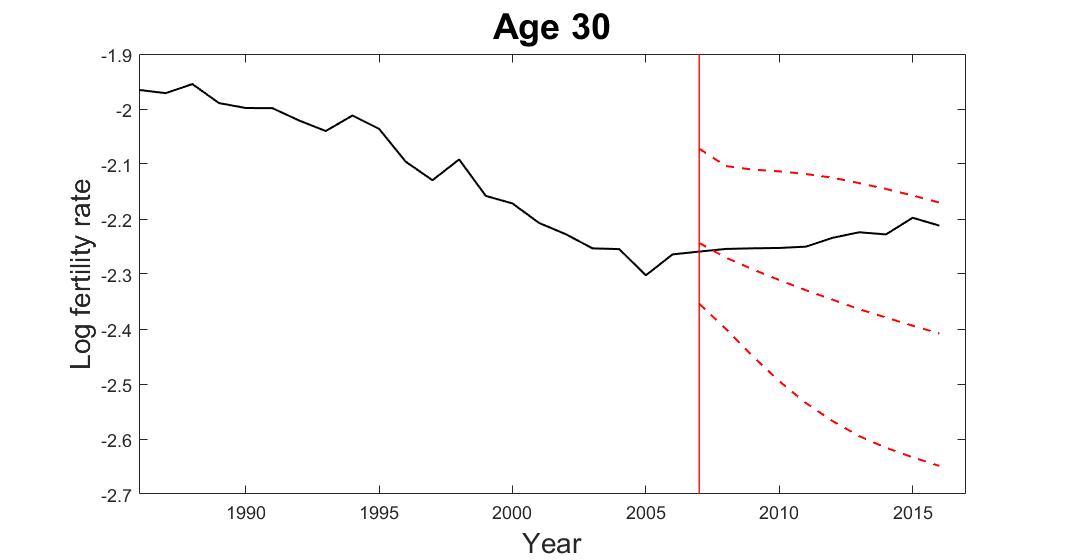}
	\end{minipage} 
	\begin{minipage}{0.5\textwidth}
		\centering
		\includegraphics[width=1\linewidth]{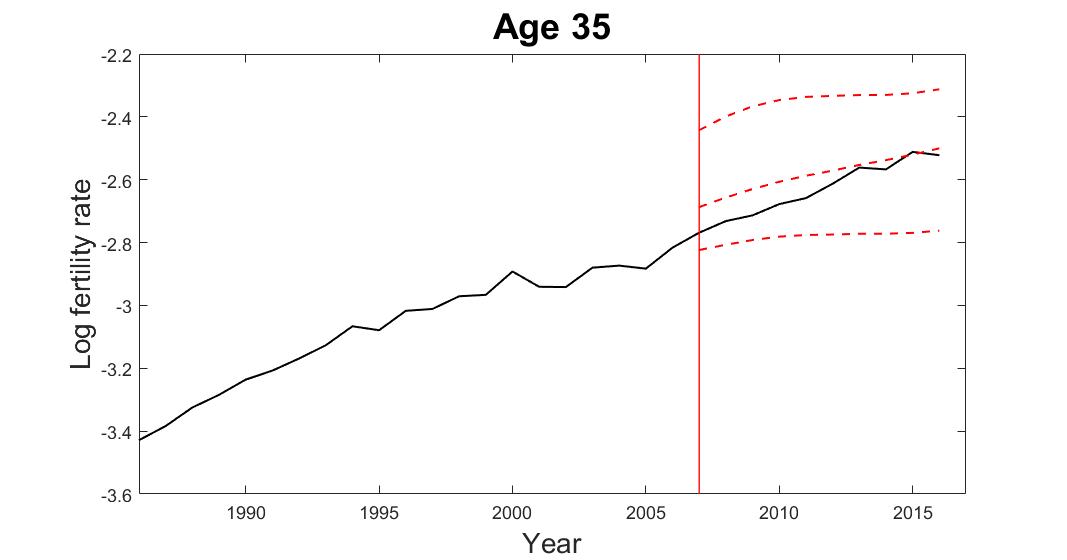}
	\end{minipage}
	\begin{minipage}{0.5\textwidth}
		\centering
		\includegraphics[width=1\linewidth]{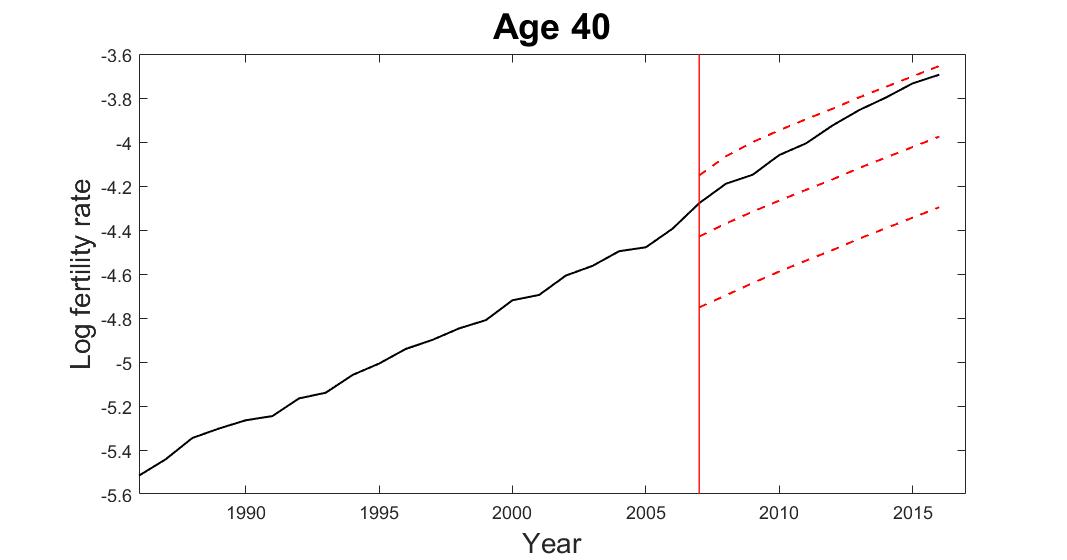}
	\end{minipage}
	\par 
	\begin{subfigure}{\linewidth}
		\centering
		\includegraphics[width=0.5\linewidth]{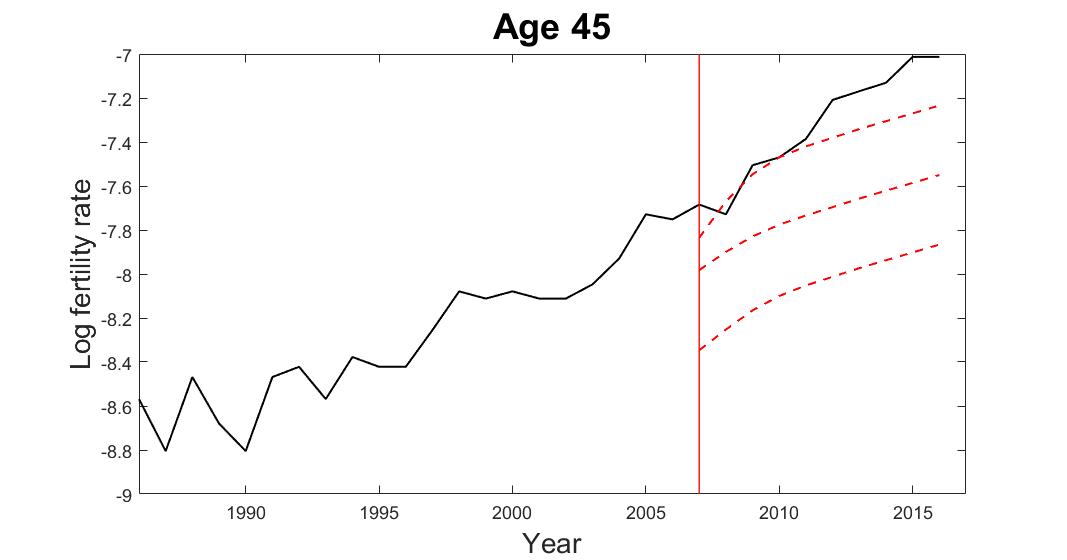}
	\end{subfigure}
	\caption{Predicted fertility rates of the selected age groups from age 15 to age 45 with 5-year age intervals using the GPR model from the year 2007 to the year 2016 based on the observations from the year 1947 to the year 2006 in Japan. The solid lines are the observed values, and the dashed lines are the predictive mean and the 95\% prediction intervals. The vertical line indicates the starting point of the predictions.}
	\label{fig: Predicted fertility rates of selected age groups from age 15 to age 45 with 5-year age intervals}
\end{figure}

\begin{figure}[!thb]
	\centering
	\includegraphics[width=1\linewidth]{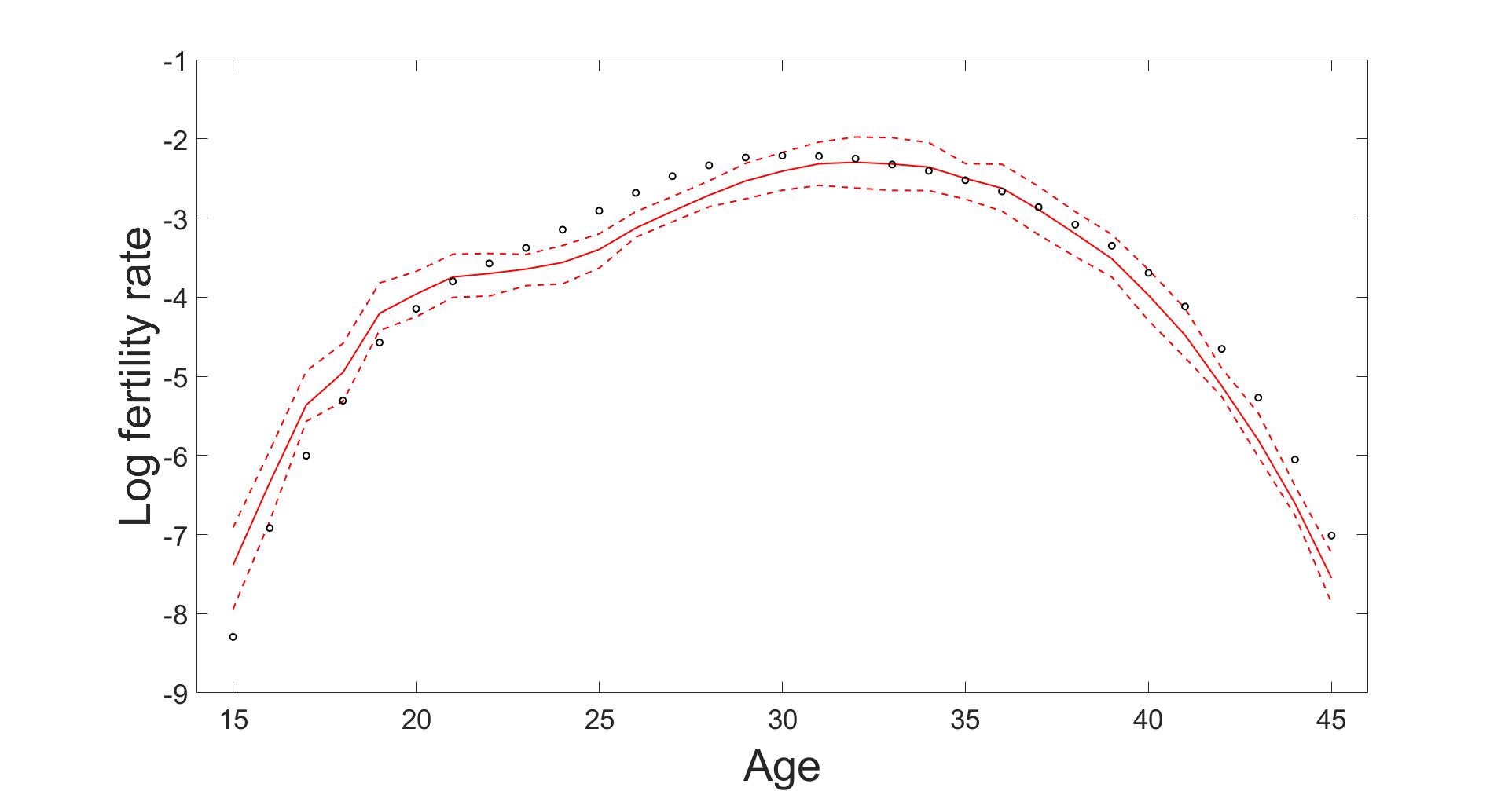}
	\caption{Predicted fertility curve from age 15 to age 45 with 95\% prediction intervals using the GPR model for the year 2016 based on the observations from the year 1947 to the year 2006 in Japan. The circles are the true log fertility rates, the solid line is the predictive mean, and the dashed lines are the 95\% prediction intervals.}
	\label{fig: Predicted fertility curve}
\end{figure}

 For the task of fertility forecasting, we again attempt to make  10-years-ahead out-of-sample forecasts of fertility rates for demonstration in this section. We maintain the same settings in the GPR model as for the mortality case, including the split of dataset, the number of interior knots in the mean structure and the Gaussian mixture components in the covariance structure with the same estimation procedures for all (hyper-) parameters involved. Figure \ref{fig: Predicted fertility rates of selected age groups from age 15 to age 45 with 5-year age intervals} presents the forecast results of the selected age-specific fertility rates by the  GPR model. We can see that the GPR method can catch the varying patterns of fertility reasonably well. Figure \ref{fig: Predicted fertility curve} shows the predicted fertility curve from age 15 to age 45 with the 95\% prediction intervals using the proposed GPR model for the year 2016 based on the observations from the year 1947 to the year 2006.

\subsection{Comparisons and forecast accuracy evaluations with existing models} \label{sec: Comparisons and forecast accuracy evaluations with existing models}
We now compare and evaluate the forecast performance and accuracy of the proposed GPR model with four other mainstream approaches in the demographic modelling literature. These include the {Lee-Carter (LC) model} \citep{lee1992modeling}, the {Lee-Miller (LM) model} \citep{lee2001evaluating}, the {Booth-Maindonald-Smith (BMS) model} \citep{booth2002applying} and the {Hyndman-Ullah (HU) model} \citep{hyndman2007robust}. \par
The LC model applies principal component analysis to decompose the age-time matrix of the log mortality (or fertility) rates $y_{x,t_{i}}$ into a linear combination of age and time parameters from the first-order principal component, i.e.
\begin{equation}\label{eq: Lee-Carter Model equation}
	y_{x,t_{i}} = a_{x} + b_{x}k_{t_{i}} + \epsilon_{x,t_{i}}, \hspace{0.5cm} \epsilon_{x,t_{i}} \sim \mathcal{N}(0, \sigma^{2}),
\end{equation}
where $a_{x}$ is the averaged log mortality (fertility) rates at age $x$ across all calender years, $b_{x}$ reflects the relative change in the log mortality (fertility) rates at age $x$, $k_{t_{i}}$ measures the general time trend of the log mortality (fertility) rates, and $\{\epsilon_{x,t_{i}}\}_{i=1}^{n}$ are the i.i.d. normally distributed error terms with zero mean and constant variance $\sigma^{2}$. The LC model relies on the $h$-step ahead extrapolated time parameter $\hat{k}_{t_{i+h}}$ by some time-series models, such as ARIMA model, to produce a $h$-step ahead forecast of the mortality (or fertility) curve across all age groups.
\par The LM model, the BMS model and the HU model can be thought of as the extensions and variants using the similar framework of the LC model in Equation (\ref{eq: Lee-Carter Model equation}). Their ways to make forecasts of the mortality (or fertility) curves also depend on the extrapolation of the time parameters derived from the age-time matrix of the log mortality (or fertility) rates \citep{booth2008mortality}. More specifically, the LM model is a modification to the LC model where the coefficient series is adjusted so that the fitted life expectancy is equal to the observed life expectancy in each year in an attempt to reduce forecast basis. The BMS model modifies the LC model to adjust the coefficients using the age-at-death distribution and determines the optimal fitting period beforehand to address the non-linearity problem in the time component. The HU model extends the LC model by adapting the functional data paradigm using non-parametric smoothing to reduce outliers in observed data and more principal components for robust forecasts.
\par
\begin{figure}[!thb]
	\centering
	\includegraphics[width=1\linewidth]{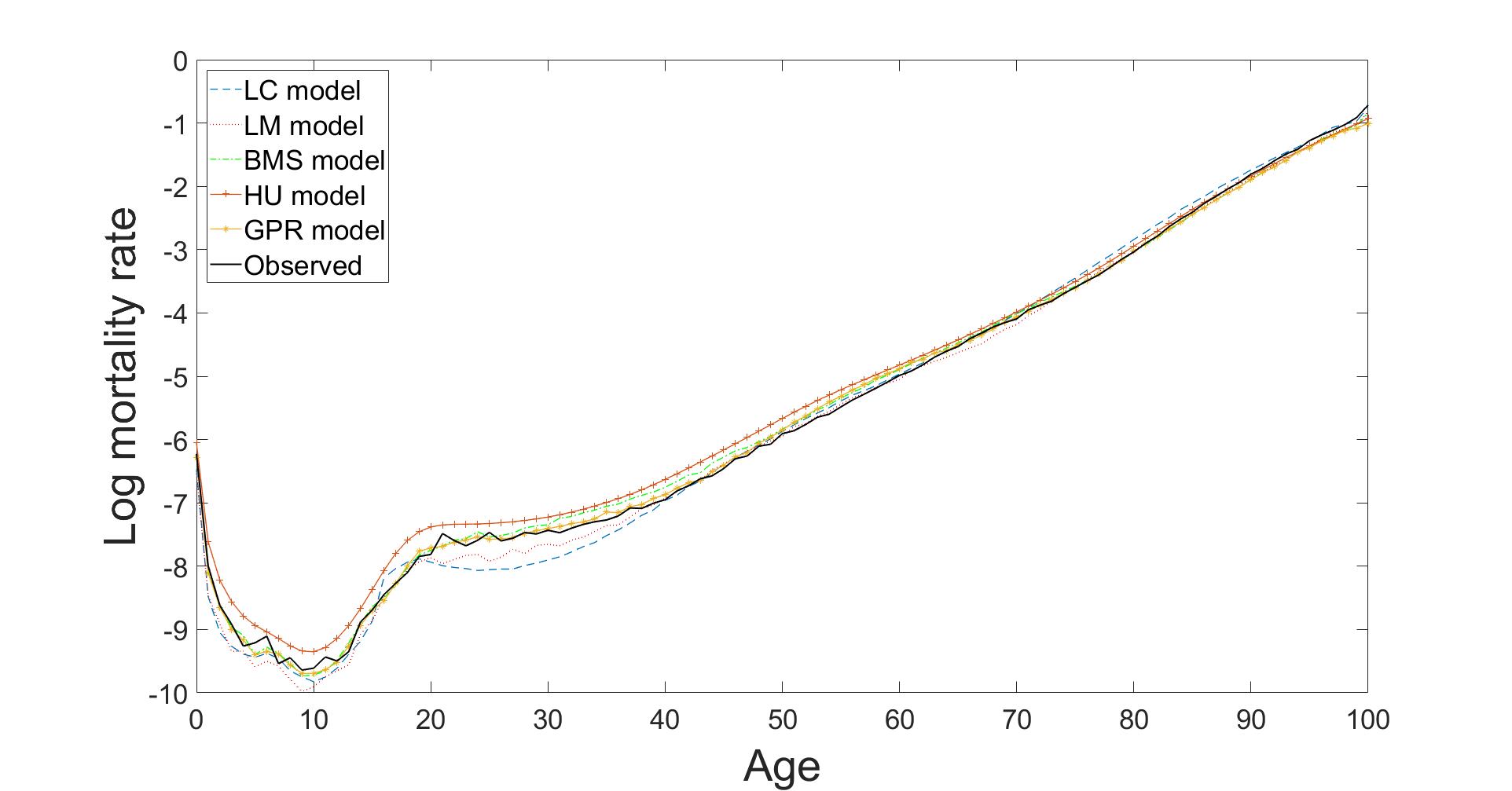}
	\caption{Predicted male mortality curves from age 0 to age 100 for the year 2016 using the LC model (with RMSE = 0.2172), the LM model (with RMSE = 0.1660), the BMS model (with RMSE = 0.1109), the HU model (with RMSE = 0.2342) and the GPR model (with RMSE = 0.0895) based on the observations from the year 1947 to the year 2006 in Japan.}
	\label{fig: Mortality comparsions with other methods}
\end{figure}

 In contrast to the LC method and its numerous variants and extensions which rely on the time parameters to decide the movement of the whole mortality (or fertility) curve across all ages groups shifting from one year to another, our approach provides a novel point of view in demographic modelling and forecasting. The proposed GPR method treats demographic data in each age group as time-series data and assumes each of them following a Gaussian process to achieve the same tasks in a discrete but intensive fashion. The natural cubic spline mean function in the proposed GPR method can automatically discover and extract more recent information from the observed data and then extrapolate its future trend in an appropriate direction. The spectral mixture covariance function in the GPR model, on the other hand, addresses the problems associated with the non-linearity and periodicity in the demographic data. Figure (\ref{fig: Mortality comparsions with other methods}) gives an example of the 10-years-ahead out-of-sample forecast results of the Japanese male mortality curves from age 0 to age 100 for the year 2016 using the LC model (with RMSE = 0.2172), the LM model (with RMSE = 0.1660), the BMS model (with RMSE = 0.1109), the HU model (with RMSE = 0.2342) and the GPR model (with RMSE = 0.0895) based on the observations from the year 1947 to the year 2006. The root mean square error (RMSE) measures the standard deviation of the average squared prediction error regardless of the positive or negative sign, and is defined here as
\[
\text{RMSE} = \sqrt{ \frac{1}{101} \sum_{x=0}^{100} \bigg( {y}_{x, t_{2016}} - \hat{y}_{x, t_{2016}} \bigg)^{2}},
\]  
where ${y}_{x, t_{2016}}$ is the log male mortality rates aged $x$ in the year 2016. \par
\begin{figure}[!thb]
	\centering
	\includegraphics[width=1\linewidth]{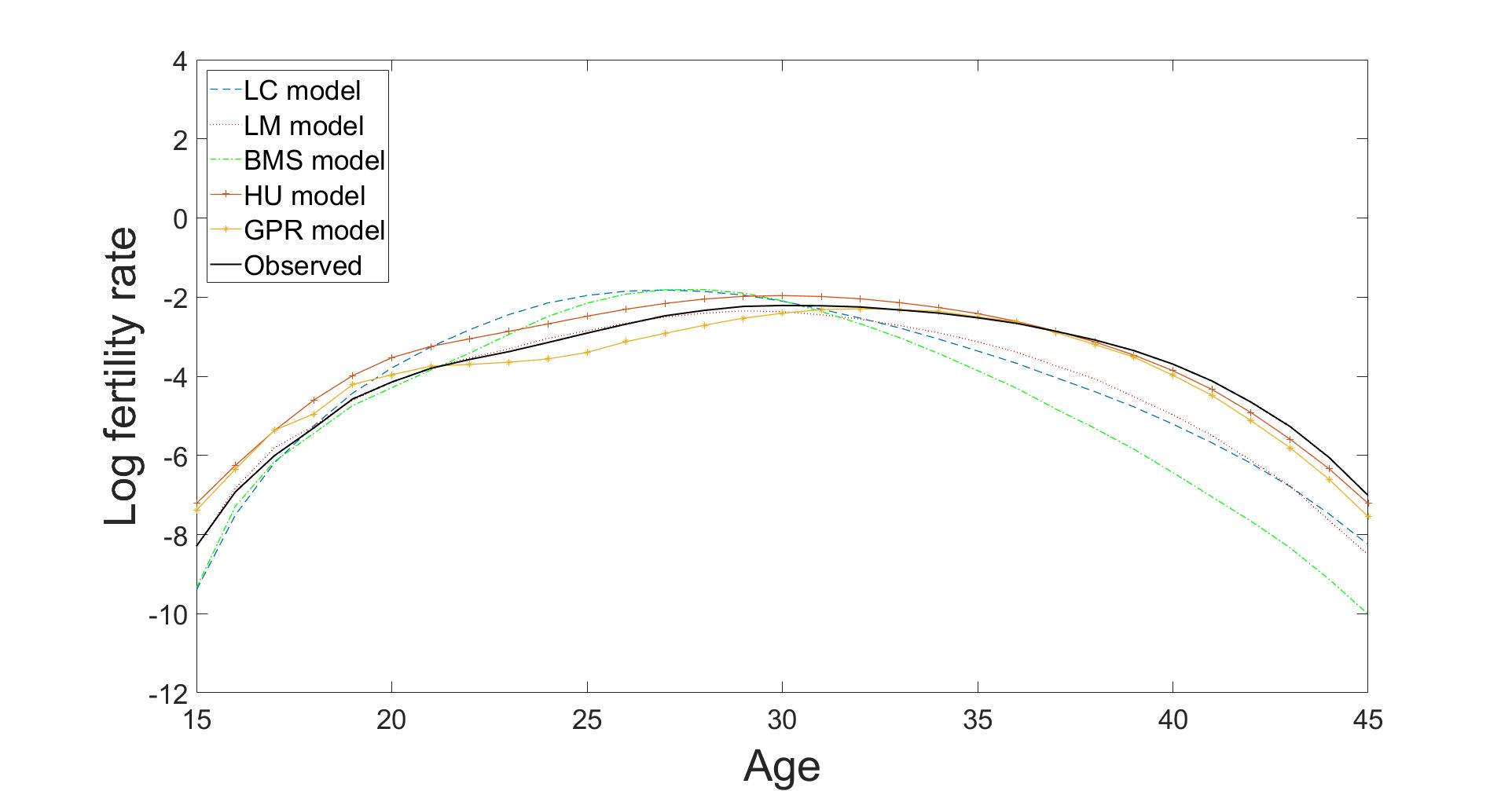}
	\caption{Predicted fertility curves from age 15 to age 45 for the year 2016 using the LC model (with RMSE = 0.9401), the LM model (with RMSE = 0.7490), the BMS model (with RMSE = 1.2198), the HU model (with RMSE = 0.4205) and the GPR model (with RMSE = 0.3764) based on the observations from the year 1947 to the year 2006 in Japan.}
	\label{fig: Fertility comparsions with other methods}
\end{figure}

 Figure (\ref{fig: Fertility comparsions with other methods}) shows another example of the 10-years-ahead out-of-sample forecast results of the Japanese fertility curves from age 15 to age 45 for the year 2016 using the LC model (with RMSE = 0.9401), the LM model (with RMSE = 0.7490), the BMS model (with RMSE = 1.2198), the HU model (with RMSE = 0.4205) and the GPR model (with RMSE = 0.3764) based on the observations from the year 1947 to the year 2006. The RMSE here is
\[
\text{RMSE} = \sqrt{ \frac{1}{31} \sum_{x=15}^{45} \bigg( {y}_{x, t_{2016}} - \hat{y}_{x, t_{2016}} \bigg)^{2}},
\]  
where ${y}_{x, t_{2016}}$ is the log fertility rates aged $x$ in the year 2016.

\subsubsection{Forecast accuracy evaluations using rolling-window analysis}
\par For evaluating the forecast accuracy, we consider ten developed countries for which mortality and fertility data are also available in the \citet{human2017university} and the \citet{humanfertility2017university}. We focus the data periods of all selected countries commencing in the year 1947 up to the year 2016 (70 years in total) for a unified purpose. Although we intent to use the same set of ten countries for both experiments in mortality and fertility, we are restricted by the availability of either mortality or fertility data in some countries. Hence, Belgium and the Netherlands are used for the mortality experiments only while Germany and Italy are selected for the fertility experiments. The remaining eight other countries are the same in both mortality and fertility experiments. We implement the four existing models mentioned above using the R package `$demography$' with the instructions of \citet{booth2014prospective} on how the parameters should be set in these models. Rolling-window analysis is used for assessing the consistency of the forecasting ability of a model by rolling a fixed size prediction interval (window) throughout the observed period \citep{zivot2007modeling}. We hold the sample data from the initial year up to the year $t_{m}$, where $t_{m} < t_{n}$, as holdout samples and produce the forecast for the $t_{m + h}$ year where $h$ is the forecast horizon. The forecasts errors are then determined by comparing the out-of-sample forecast result with the actual data. We increase one rolling-window (1 year ahead) in year $t_{m + 1}$ to make the same procedure again for the year $t_{m + 1 + h}$ until the rolling-window analysis covers all available data in year $t_{n}$. We consider four different forecast horizons $(h = 5, 10, 15$ and $20)$ with ten sets of rolling-window to examine the short-term, the mid-term, and the long-term forecast abilities of the models. In the mortality rolling-window experiments, the RMSE formula is 
\[
\text{RMSE}_{c}(h) = \sqrt{\frac{1}{10\times 101}\sum_{w=0}^{9}\sum_{x=0}^{100} \bigg( y_{x, t_{m+w+h}} - \hat{y}_{x, t_{m+w+h}} \bigg)^{2}},
\]   
where $c$ is the selected country, $w$ is the index of the rolling-window sets and $x$ is the age from age 0 to age 100. And the RMSE in the fertility rolling-window experiments is \[
\text{RMSE}_{c}(h) = \sqrt{\frac{1}{10\times 31}\sum_{w=0}^{9}\sum_{x=15}^{45} \bigg( y_{x,t_{m + w + h}} - \hat{y}_{x,t_{m + w + h}} \bigg)^{2}},
\]   
where $x$ is the age from age 15 to age 45.
\par Table \ref{table: Forecast accuracy of male mortality by the average RMSEs in ten sets of rolling-window analysis} presents the average RMSE results of ten sets of rolling-window analysis across ten countries in four different forecast horizons for the mortality  experiments. The proposed GPR model performs consistently the most desirable for mortality forecasting. It occupies the major positions of having the least forecast errors among the ten selected countries in the four different prediction horizons. The method is shown to be capable of capturing various mortality curve patterns across different periods and age groups. We can see that the GPR model has significant improvements in the prediction accuracy in comparison with the four other tested models, particularly for the mortality data of Japan. It may mainly be thanks to the intensive treatment of mortality curve fitting by the GPR approach, which enhances the preciseness and robustness, especially for countries with high mortality fluctuations in certain age groups due possibly to natural disasters. It is also worth noting that the forecasting performances of the LM model in the short term and the mid-short term are remarkably well with the UK and the USA male mortality data. The consistency and small variabilities of mortality curves in the UK and the USA over age and time may contribute to the superiority of the LM model compared to the other tested models in our experiments. The forecast performances of the LC model, the BMS model and the HU model are reasonably similar with no particularly outstanding point in our experiments.
\par Table \ref{table: Forecast accuracy of fertility curves by the average RMSEs in ten sets of rolling-window analysis} shows the average RMSE results of ten sets of rolling-window analysis across the selected ten countries in the four different prediction horizons in the fertility  experiments. The proposed GPR model continues to maintain the dominant positions of having the smallest forecast errors in the short term and the mid-short term prediction horizons and is equally well with the HU model in the fifteen-year forecast horizon. As for the long term prediction of the twenty-year forecast horizon, it seems that the HU model is more suitable to capture the fertility patterns with smaller forecasts errors than the GPR model. The HU model also fits the French fertility data better than the other models. It may be because the fertility data contain more outliers or measurement errors than the mortality data, and the HU model includes the smoothing techniques, which can improve the model forecasting performances if the observed fertility data are smoothed in advance.

\par 
\begin{table}[ht]
	\centering
	\begin{tabular}{cccccccccccccccccccc}  
		\toprule
		\multirow{1}{*}{
			\parbox[c]{.1\linewidth}{\centering Country}}
		& \multicolumn{3}{c}{LC model} &&
		\multicolumn{3}{c}{LM model} &&
		\multicolumn{3}{c}{BMS model} && 
		\multicolumn{3}{c}{HU model} &&
		\multicolumn{3}{c}{GPR model}\\ 
		\midrule
		\underline{$h = 5$}\\
		Austria &&  {\centering0.3122} && &&  {\centering0.3191
		}  && &&  {\centering0.2596} && &&  {\centering0.2591}  && &&  \cellcolor{yellow}{\centering0.2581} \\
		
		Belgium &&  {\centering0.2560} && &&  {\centering0.2783
		}  && &&  {\centering0.2429} && && {\centering0.2458}  && &&  \cellcolor{yellow}{\centering0.2359} \\
		
		Canada &&  {\centering0.1850} && &&  {\centering0.1574}  && &&  {\centering0.1762} && &&  \cellcolor{yellow}{\centering0.1519}  && &&  {\centering0.1583} \\
		
		France &&  {\centering0.1879}&&&&  \cellcolor{yellow}{\centering0.1260}&&&&  
		{\centering0.1365} &&&&  {\centering0.1477} &&&&  {\centering0.1457} \\
		
		Japan &&  {\centering0.1903} && &&  {\centering0.1365}  && &&  {\centering0.1564} && &&  {\centering0.1533}  && &&  \cellcolor{yellow}{\centering0.1164} \\
		
		Netherlands &&  {\centering0.2679
		} && &&  {\centering0.2284}  && &&  \cellcolor{yellow}{\centering0.2046} && &&  {\centering0.2265}  && &&  {\centering0.2233} \\
		
		Sweden &&  {\centering0.2873} && &&  {\centering0.3178}  && &&  {\centering0.2695} && &&  {\centering0.2633}  && &&  \cellcolor{yellow}{\centering0.2566} \\
		
		Switzerland &&  {\centering0.3346} && &&  {\centering0.3597
		}  && &&  {\centering0.3077} && &&  {\centering0.3109}  && &&  \cellcolor{yellow}{\centering0.2774} \\
		
		UK &&  {\centering0.1805} && &&  \cellcolor{yellow}{\centering0.1301}  && &&  {\centering0.1382} && &&  {\centering0.1584}  && &&  {\centering0.1504} \\
		
		USA &&  {\centering0.1266} && &&  \cellcolor{yellow}{\centering0.0875}  && &&  {\centering0.1324} && &&  {\centering0.1183}  && &&  {\centering0.1258} \\
		
		{Average} &&  {\centering0.2328} && &&  {\centering0.2141}  && &&  {\centering0.2024} && &&  {\centering0.2035}  && &&  \cellcolor{yellow}{\centering{0.1948}} \\
		\midrule
		\underline{$h = 10$}\\
		
		Austria &&  {\centering0.3504} && &&  {\centering0.3289}  && &&  {\centering0.2793} && &&  {\centering0.2809}  && &&  \cellcolor{yellow}{\centering0.2783} \\
		
		Belgium &&  {\centering0.2902} &&&&  {\centering0.2945}  && &&  {\centering0.2820} && &&  {\centering0.3044}  && &&  \cellcolor{yellow}{\centering0.2634} \\
		
		Canada &&  {\centering0.2197} && &&  \cellcolor{yellow}{\centering0.1837}  && &&  {\centering0.2078} && &&  {\centering0.1862}  && &&  {\centering0.2017} \\
		
		France &&  {\centering0.2443} && &&  \cellcolor{yellow}{\centering0.1871}  && &&  {\centering0.2074} && &&  {\centering0.2391}  && &&  {\centering0.1985} \\
		
		Japan &&  {\centering0.2630} && &&  {\centering0.2187}  && &&  {\centering0.2824} && &&  {\centering0.2382}  && &&  \cellcolor{yellow}{\centering0.1208} \\
		
		Netherlands &&  {\centering0.3294
		} && &&  {\centering0.2706}  && &&  \cellcolor{yellow}{\centering0.2629} && &&  {\centering0.2861}  && &&  {\centering0.2678} \\
		
		Sweden &&  {\centering0.3153} && &&  {\centering0.3311
		}  && &&  {\centering0.2869} && && {\centering0.2813}  && &&   \cellcolor{yellow}{\centering0.2740} \\
		
		Switzerland &&  {\centering0.3888
		} && &&  {\centering0.4087
		}  && &&  {\centering0.3942} && &&  {\centering0.4030}  && &&  \cellcolor{yellow}{\centering0.3842} \\
		
		UK &&  {\centering0.2271} && &&  \cellcolor{yellow}{\centering0.1730} && &&  {\centering0.1987} && &&  {\centering0.2133}  && &&  {\centering0.1899} \\
		
		USA &&  {\centering0.1514} && &&  \cellcolor{yellow}{\centering0.1231}  && &&  {\centering0.1706} && &&  {\centering0.1474}  && &&  {\centering0.1724} \\
		{Average} &&  {\centering0.2780
		} && &&  {\centering0.2519
		}  && &&  {\centering0.2572
		} && &&  {\centering0.2580
		}  && &&  \cellcolor{yellow}{\centering{0.2351}} \\
		\bottomrule
	\end{tabular}
	\caption{The average RMSEs of ten sets of rolling-windows analysis on the predicted male mortality curves using the LC model, the LM model, the BMS model, the HU model and the GPR model. The lowest forecast errors among all the models are highlighted.} 
	\label{table: Forecast accuracy of male mortality by the average RMSEs in ten sets of rolling-window analysis}
\end{table}

\begin{table}[ht]
	\ContinuedFloat
	\centering
	\begin{tabular}{cccccccccccccccccccc}  
		\toprule
		\multirow{1}{*}{
			\parbox[c]{.1\linewidth}{\centering Country}}
		& \multicolumn{3}{c}{LC model} &&
		\multicolumn{3}{c}{LM model} &&
		\multicolumn{3}{c}{BMS model} && 
		\multicolumn{3}{c}{HU model} &&
		\multicolumn{3}{c}{GPR model}\\ 
		\midrule
		\underline{$h = 15$}\\
		
		Austria &&  {\centering0.4022} && &&  {\centering0.3707} && &&  \cellcolor{yellow}{\centering0.3119} && &&  {\centering0.3582}  && &&  {\centering0.3130} \\
		
		Belgium &&  {\centering0.3251} && &&  {\centering0.3148}  && &&  {\centering0.3050} && &&  {\centering0.3558}  && &&  \cellcolor{yellow}{\centering0.2906} \\
		
		Canada &&  {\centering0.2659
		} && &&  {\centering0.2413}  && &&  {\centering0.2340} && &&  \cellcolor{yellow}{\centering0.2220}  && &&  {\centering0.2595} \\
		
		France &&  {\centering0.2992} && &&  \cellcolor{yellow}{\centering0.2699}  && &&  {\centering0.2937} && &&  {\centering0.3431}  && &&  {\centering0.2924} \\
		
		Japan &&  {\centering0.3631} && &&  {\centering0.3366}  && &&  {\centering0.3551} && &&  {\centering0.2688}  && &&  \cellcolor{yellow}{\centering0.2063} \\
		
		Netherlands &&  {\centering0.3787} && &&  {\centering0.3316}  && &&  {\centering0.3145} && &&  {\centering0.3335}  && &&  \cellcolor{yellow}{\centering0.2992} \\
		
		Sweden &&  {\centering0.3677} && &&  {\centering0.3637}  && &&  {\centering0.3194} && &&  {\centering0.3331}  && &&  \cellcolor{yellow}{\centering0.2986} \\
		
		Switzerland &&  \cellcolor{yellow}{\centering0.4574} && &&  {\centering0.4724}  && &&  {\centering0.5270} && &&  {\centering0.5347}  && &&  {\centering0.5279} \\
		
		UK &&  {\centering0.2836} && &&  \cellcolor{yellow}{\centering0.2292}  && &&  {\centering0.2373} && &&  {\centering0.2604}  && &&  {\centering0.2310} \\
		
		USA &&  {\centering0.1855} && &&  {\centering0.1825}  && &&  {\centering0.2109} && &&  \cellcolor{yellow}{\centering0.1798}  && &&  {\centering0.2172} \\
		{Average} &&  {\centering0.3328
		} && &&  {\centering0.3113
		}  && &&  {\centering0.3109} && &&  {\centering0.3189}  && &&  \cellcolor{yellow}{\centering{0.2936}} \\
		\midrule
		\underline{$h = 20$}\\
		
		Austria &&  {\centering0.4600
		} && &&  {\centering0.4289
		}  && &&  {\centering0.3833
		} && &&  {\centering0.4275
		}  && &&  \cellcolor{yellow}{\centering0.3813} \\
		
		Belgium &&  {\centering0.3666} && &&  {\centering0.3402
		}  && &&  {\centering0.3330} && &&  {\centering0.3740
		}  && && 
		\cellcolor{yellow}{\centering0.3200} \\
		
		Canada &&  {\centering0.3113
		} && &&  {\centering0.2747
		}  && &&  \cellcolor{yellow}{0.2394} && &&  {\centering0.2594
		}  && &&  {\centering0.2613} \\
		
		France &&  {\centering0.3491} && &&  {\centering0.3218}  && &&  {\centering0.3236} && &&  {\centering0.4051}  && &&  \cellcolor{yellow}{\centering0.3105} \\
		
		Japan &&  {\centering0.5193} && &&  {\centering0.5120}  && &&  {\centering0.4385} && &&  {\centering0.3333}  && &&  \cellcolor{yellow}{\centering0.2828} \\
		
		Netherlands &&  {\centering0.4218} && &&  {\centering0.3828}  && &&  {\centering0.3968} && &&  {\centering0.3739}  && &&  \cellcolor{yellow}{\centering0.3321} \\
		
		Sweden &&  {\centering0.4214} && &&  {\centering0.3943}  && &&  {\centering0.3794} && &&  {\centering0.4154}  && &&  \cellcolor{yellow}{\centering0.3270} \\
		
		Switzerland &&  \cellcolor{yellow}{\centering0.4910} && &&  {\centering0.5259}  && &&  {\centering0.5930} && &&  {\centering0.6774}  && &&  {\centering0.6169} \\
		
		UK &&  {\centering0.3529} && &&  {\centering0.2894}  && &&  {\centering0.2750} && &&  {\centering0.3774}  && &&  \cellcolor{yellow}{\centering0.2717} \\
		
		USA &&  {\centering0.2017
		} && &&  {\centering0.2089}  && &&  \cellcolor{yellow}{\centering0.1856} && &&  {\centering0.2159}  && &&  {\centering0.2122} \\
		
		{Average} &&  {\centering0.3895} && &&  {\centering0.3679} && &&  {\centering0.3548} && &&  {\centering0.3859}  && &&  \cellcolor{yellow}{\centering{0.3316}} \\
		\bottomrule
	\end{tabular}
	\caption{(\textit{cont.}) The average RMSEs of ten sets of rolling-windows analysis on the predicted male mortality curves using the LC model, the LM model, the BMS model, the HU model and the GPR model. The lowest forecast errors among all the models are highlighted.} 
\end{table}

\begin{table}[ht]
	\centering
	\begin{tabular}{cccccccccccccccccccc}  
		\toprule
		\multirow{1}{*}{
			\parbox[c]{.1\linewidth}{\centering Country}}
		& \multicolumn{3}{c}{LC model} &&
		\multicolumn{3}{c}{LM model} &&
		\multicolumn{3}{c}{BMS model} && 
		\multicolumn{3}{c}{HU model} &&
		\multicolumn{3}{c}{GPR model}\\ 
		\midrule
		\underline{$h = 5$}\\
		
		Austria &&  {\centering0.6112
		} && &&  {\centering0.2509
		}  && &&  {\centering0.6203
		} && &&  {\centering0.2287
		}  && &&  \cellcolor{yellow}{\centering0.1801
		} \\
		
		Canada &&  {\centering0.6044
		} && &&  {\centering0.2481
		}  && &&  \cellcolor{yellow}{\centering0.1330} && &&  {\centering0.3236}  && &&  {\centering0.2432} \\
		
		France &&  {\centering0.4927} && &&  {\centering0.1565}  && &&  {\centering0.2334} && &&  \cellcolor{yellow}{\centering0.0981}  && &&  {\centering0.2462} \\
		
		Germany &&  {\centering0.6495} && &&  {\centering0.2420}  && &&  {\centering0.5870
		} && &&  {\centering0.1608}  && &&  \cellcolor{yellow}{\centering0.1489} \\	
		
		Italy &&  {\centering0.5414} && &&  \cellcolor{yellow}{\centering0.2635}  && &&  {\centering0.3719} && &&  {\centering0.4402}  && &&  {\centering0.2805} \\		
		
		Japan &&  {\centering0.6533} && &&  {\centering0.3876}  && &&  {\centering0.6721} && &&  {\centering0.5302}  && &&  \cellcolor{yellow}{\centering0.2769} \\
		
		Sweden &&  {\centering0.5970} && &&  {\centering0.1905}  && &&  {\centering0.1562} && &&  {\centering0.2768}  && &&  \cellcolor{yellow}{\centering0.1553} \\
		
		Switzerland &&  {\centering0.6734} && &&  {\centering0.2822}  && &&  {\centering0.4553} && &&  {\centering0.2536}  && &&  \cellcolor{yellow}{\centering0.2059} \\
		
		UK &&  {\centering0.4280} && &&  \cellcolor{yellow}{\centering0.2392}  && &&  {\centering0.2554} && &&  {\centering0.2970}  && &&  {\centering0.2920} \\
		
		USA &&  {\centering0.4499} && &&  \cellcolor{yellow}{\centering0.1757}  && &&  {\centering0.2170} && &&  {\centering0.3554}  && &&  {\centering0.2431} \\
		{Average} &&  {\centering0.5701} && &&  {\centering0.2436}  && &&  {\centering0.3702} && &&  {\centering0.2964}  && &&  \cellcolor{yellow}{\centering{0.2272}} \\
		\midrule
		\underline{$h = 10$}\\
		
		Austria &&  {\centering0.7328} && &&  {\centering0.5081}  && &&  {\centering0.7915} && &&  {\centering0.3370}  && &&  \cellcolor{yellow}{\centering0.3276} \\
		
		Canada &&  {\centering0.7476} && &&  {\centering0.5054}  && &&  \cellcolor{yellow}{\centering0.2581} && &&  {\centering0.3630}  && &&  {\centering0.2881} \\
		
		France &&  {\centering0.6026} && &&  {\centering0.3428}  && &&  {\centering0.4888} && &&  \cellcolor{yellow}{\centering0.2043}  && &&  {\centering0.3742} \\
		
		Germany &&  {\centering0.7531} && &&  {\centering0.4990}  && &&  {\centering0.7365
		} && &&  {\centering0.2909}  && &&  \cellcolor{yellow}{\centering0.2008} \\	
		
		Italy &&  {\centering0.7064} && &&  {\centering0.5449}  && &&  {\centering0.8189} && &&  {\centering0.5346}  && &&  \cellcolor{yellow}{\centering0.4802} \\
		
		Japan &&  {\centering0.8998
		} && &&  {\centering0.7846}  && &&  {\centering1.2119} && &&  {\centering0.6443}  && &&  \cellcolor{yellow}{\centering0.3424} \\
		
		Sweden &&  {\centering0.7524} && &&  {\centering0.4169}  && &&  {\centering0.2699} && &&  {\centering0.3857}  && &&  \cellcolor{yellow}{\centering0.1789} \\
		
		Switzerland &&  {\centering0.7926} && &&  {\centering0.5536}  && &&  {\centering0.4835} && &&  {\centering0.4489}  && &&  \cellcolor{yellow}{\centering0.3274} \\
		
		UK &&  {\centering0.5928} && &&  {\centering0.4780}  && &&  {\centering0.3448} && &&  {\centering0.3594}  && &&  \cellcolor{yellow}{\centering0.3294} \\
		
		USA &&  {\centering0.5058} && &&  {\centering0.3172}  && &&  \cellcolor{yellow}{\centering0.3092} && &&  {\centering0.4458}  && &&  {\centering0.4241} \\
		
		{Average} &&  {\centering0.7086} && &&  {\centering0.4950}  && &&  {\centering0.5713} && &&  {\centering0.4014}  && &&  \cellcolor{yellow}{\centering{0.3273}} \\
		\bottomrule
	\end{tabular}
	\caption{The average RMSEs of ten sets of rolling-windows analysis on the predicted fertility curves using the LC model, the LM model, the BMS model, the HU model and the GPR model. The lowest forecast errors among all the models are highlighted.} 
	\label{table: Forecast accuracy of fertility curves by the average RMSEs in ten sets of rolling-window analysis}
\end{table}

\begin{table}[ht]
	\ContinuedFloat
	\centering
	\begin{tabular}{cccccccccccccccccccc}  
		\toprule
		\multirow{1}{*}{
			\parbox[c]{.1\linewidth}{\centering Country}}
		& \multicolumn{3}{c}{LC model} &&
		\multicolumn{3}{c}{LM model} &&
		\multicolumn{3}{c}{BMS model} && 
		\multicolumn{3}{c}{HU model} &&
		\multicolumn{3}{c}{GPR model}\\ 
		\midrule
		\underline{$h = 15$}\\
		
		Austria &&  {\centering0.8863} && &&  {\centering0.7721}  && &&  {\centering0.9997} && &&  \cellcolor{yellow}{\centering0.4972}  && &&  {\centering0.6215} \\
		
		Canada &&  {\centering0.9300} && &&  {\centering0.8080}  && &&  {\centering0.5972} && && \cellcolor{yellow}{\centering0.4347}  && &&  {\centering0.4894} \\
		
		France &&  {\centering0.7871} && &&  {\centering0.5908}  && &&  {\centering0.8119} && &&  \cellcolor{yellow}{\centering0.4960}  && &&  {\centering0.5973} \\
		
		Germany &&  {\centering0.9231} && &&  {\centering0.7684}  && &&  {\centering0.9732} && &&  {\centering0.4319}  && &&  \cellcolor{yellow}{\centering0.2961} \\	
		
		Italy &&  {\centering0.9339} && &&  {\centering0.8319}  && &&  {\centering1.3051} && &&  \cellcolor{yellow}{\centering0.7208}  && &&  {\centering0.9365} \\		
		
		Japan &&  {\centering1.2392} && &&  {\centering1.2202}  && &&  {\centering1.5252} && &&  {\centering0.6989}  && &&  \cellcolor{yellow}{\centering0.5641} \\
		
		Sweden &&  {\centering0.8930} && &&  {\centering0.6031}  && &&  {\centering0.5837} && &&  {\centering0.5307}  && &&  \cellcolor{yellow}{\centering0.3582} \\
		
		Switzerland &&  {\centering0.9837} && &&  {\centering0.8390}  && &&  {\centering0.9742} && &&  {\centering0.6711}  && &&  \cellcolor{yellow}{\centering0.6488} \\
		
		UK &&  {\centering0.7850} && &&  {\centering0.7286
		}  && && \cellcolor{yellow}{\centering0.4473} && &&  {\centering0.5277}  && && {\centering0.4505} \\
		
		USA &&  {\centering0.6125} && &&  {\centering0.5309}  && &&  \cellcolor{yellow}{\centering0.4280} && &&  {\centering0.5312}  && &&  {\centering0.5539} \\	
		{Average} &&  {\centering0.8986} && &&  {\centering0.7690}  && &&  {\centering0.8495} && &&  {\centering0.5603}  && &&  \cellcolor{yellow}{\centering{0.5439}} \\
		
		\midrule
		\underline{$h = 20$}\\
		
		Austria &&  {\centering1.0445} && &&  {\centering1.0165}  && &&  {\centering1.2876} && &&  \cellcolor{yellow}{\centering0.6219}  && &&  {\centering1.0984} \\
		
		Canada &&  {\centering1.2030} && &&  {\centering1.1651}  && &&  {\centering1.2181} && &&  \cellcolor{yellow}{\centering0.5261}  && &&  {\centering0.9405} \\
		
		France &&  {\centering0.9948} && &&  {\centering0.8514
		}  && &&  {\centering1.1597
		} && &&  \cellcolor{yellow}{\centering0.8039
		}  && &&  {\centering0.9926
		} \\
		
		Germany &&  {\centering1.1324} && &&  {\centering1.0523}  && &&  {\centering1.3186} && &&  {\centering0.6835}  && &&  \cellcolor{yellow}{\centering0.6710} \\	
		
		Italy &&  {\centering1.1945} && &&  {\centering1.1354}  && &&  {\centering1.6735} && &&  \cellcolor{yellow}{\centering0.9316}  && &&  {\centering1.6623} \\
		
		Japan &&  {\centering1.6871} && &&  {\centering1.7571}  && &&  {\centering1.6005} && &&  \cellcolor{yellow}{\centering0.7553}  && &&  {\centering0.8688} \\
		
		Sweden &&  {\centering1.0161} && &&  {\centering0.7615}  && &&  {\centering0.9025} && &&  {\centering0.6986}  && &&  \cellcolor{yellow}{\centering0.6580} \\
		
		Switzerland &&  {\centering1.2332} && &&  {\centering1.1462}  && &&  {\centering1.4133} && &&  \cellcolor{yellow}{\centering0.7295}  && &&  {\centering1.2452} \\
		
		UK &&  {\centering1.0005} && &&  {\centering0.9784}  && &&  {\centering0.7820} && &&  {\centering0.8654}  && &&  \cellcolor{yellow}{\centering0.6823} \\
		
		USA &&  {\centering0.7946} && &&  {\centering0.7948}  && &&  {\centering0.7017} && &&  \cellcolor{yellow}{\centering0.6077}  && &&  {\centering0.6676} \\
		{Average} &&  {\centering1.1301} && &&  {\centering1.0659}  && &&  {\centering1.2057} && &&  \cellcolor{yellow}{\centering{0.7224}}  && &&  {\centering0.9487} \\
		\bottomrule
	\end{tabular}
	\caption{(\textit{cont.}) The average RMSEs of ten sets of rolling-windows analysis on the predicted fertility curves using the LC model, the LM model, the BMS model, the HU model and the GPR model. The lowest forecast errors among all the models are highlighted.}
\end{table}
\section{Discussion and conclusion remarks}\label{sec: Discussion and conclusion remarks}
Following the theoretical framework of Gaussian process regression discussed in Section \ref{sec: GPR Theoretical background of Gaussian process regression}, in this article, we have introduced a new design of the Gaussian process regression model equipped with the natural cubic spline mean function and the spectral mixture covariance function as a new approach for mortality and fertility modelling and forecasting. The use of the natural cubic spline mean function in the proposed GPR model can exploit the local information of the recent data and force its extrapolation beyond the observed data as a linear function to provide smooth and robust forecasts. The spectral mixture covariance function, on the other hand, detects and handles any non-linearity and periodicity of the demographic data unexplained by the fitted mean function.
\par We have demonstrated the usefulness and flexibility of the proposed GPR model through two empirical data applications: one is to forecast the male mortality data, and the other is to predict the fertility data. Our experiments have proved the forecasting ability of the proposed GPR model. In these two experiments, the accuracies of the predicted mortality and fertility curves are significantly improved by the proposed GPR model in terms of the forecast errors for the ten tested countries in different forecasting horizons compared to the four other mainstream approaches. The prediction performances of these four methods rely barely on the extrapolation of time parameters as they decide how the movement of the whole mortality (or fertility) curve should shift over time across all age groups. The entire fitted curve can go far away from its expected location when the fitted time parameters in these four methods are not well predicted, and this problem can be seen in our demonstration for the predicted Japanese fertility curve in Figure \ref{fig: Fertility comparsions with other methods}. The forecast results of the fertility curves by the mainstream approaches, such as the LC model and the BMS model, deviate from the observation noticeably in some parts of the curve. In contrast, the proposed method is more robust to this issue, because it treats the demographic data in each age group as time-series data and assumes each of them following a Gaussian process to achieve the same tasks in a discrete but very intensive fashion. This issue can hence be avoided as any misprediction by the proposed model only affects a single point of the curve in each age group, rather than the shape of the whole curve across all the age groups.
\par Furthermore, the proposed GPR model is in time series, non-linear and non-parametric structures and is not restricted to mortality and fertility forecasting and modelling only. It can also support a wide range of potential applications in many domains of applied science and engineering, such as signal processing or weather forecasting. Its particular features enable more flexibility than the four existing models considered in our study. Although \cite{wu2018gaussian} proposed a similar approach using a Gaussian process regression with a weighted linear mean function, the weighted linear mean structure relies on the assumption that more recent data tend to have more weights on the future mortality, which requires on subjective judgements on how the weights are assigned to reflect the impacts of different periods. It can affect the accuracy of forecasts remarkably when different subjective choices of weights are adopted, or the weight parameters are inappropriately specified \citep{wu2018gaussian}. In contrast, our method exploits the natural cubic spline function as a non-parametric mean structure without any assumption made on the relationship between the mortality rates and the time, and it lets the natural cubic spline mean function automatically identify the local information from the observed data and then project the future mortality trend in an appropriate direction.
\par The main limitations of the proposed GPR models also relate to the characteristics of the classes of independent time series and simple non-parametric single population extrapolative methods of which it belongs to. Although it can capture the trends of the historical demographic data robustly, at the same time, it is lack of ability to incorporate and model more other related information. For example, one may expect that the mortality or fertility rates for different ages may be closely correlated, especially among the neighbouring ages. Our GPR model would be more desirable if it could model the dependence on different age groups simultaneously while taking their heterogeneity into account. We, therefore, aim to extend the current GPR model to a spatial or a multi-output GPR model for connecting the relationships between different age groups and time periods altogether. These will be left for our future works to achieve.    
\begin{spacing}{1.5}
	\bibliographystyle{apacite}
	\bibliography{reference}
\end{spacing}
\end{document}